\documentclass[10pt,twocolumn,letterpaper]{article}

\usepackage[pagenumbers]{cvpr}      % To produce the REVIEW version
\usepackage{bm}
\usepackage{amsmath}
\usepackage{multirow}

\newcommand{\red}[1]{{\color{red}#1}}

\definecolor{cvprblue}{rgb}{0.21,0.49,0.74}
\usepackage[pagebackref,breaklinks,colorlinks,allcolors=cvprblue]{hyperref}

\def\paperID{3107} % *** Enter the Paper ID here
\def\confName{CVPR}
\def\confYear{2026}

\newcommand\blfootnote[1]{%
  \begingroup
  \vspace{-0.25cm}
  \renewcommand\thefootnote{}\footnote{#1}%
  \addtocounter{footnote}{-1}%
  \endgroup
}

\title{ARGS: \red{A}uto-\red{R}egressive \red{G}aussian \red{S}platting \\ via Parallel Progressive Next-Scale Prediction}

\author{
Quanyuan Ruan$^1$\footnotemark[2] \quad
Kewei Shi$^{2}$\footnotemark[2] \quad
Jiabao Lei$^{3}$ \quad
Xifeng Gao$^4$\footnotemark[1] \quad
Xiaoguang Han$^3$\footnotemark[1] \and
$^{1}$South China University of Technology \quad
$^{2}$The University of Hong Kong \\
$^{3}$The Chinese University of Hong Kong, Shenzhen \quad
$^{4}$Lightspeed
}

\begin{document}

\twocolumn[{%
\renewcommand\twocolumn[1][]{#1}%
\maketitle
\begin{center}
    \centering
    \vspace{-0.5cm}
    \captionsetup{type=figure}
    \includegraphics[width=1.0\linewidth]{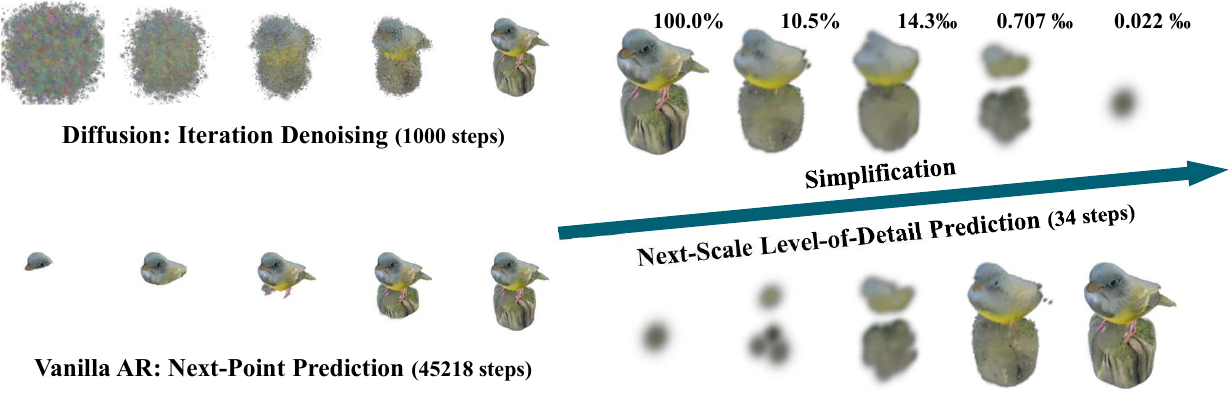}
    \captionof{figure}{We propose a novel framework for generating Gaussian Splatting fields using a next-scale autoregressive prediction paradigm. Unlike diffusion-based approaches (e.g., DDPM\cite{ho2020denoising}), which iteratively denoise latent representations, or conventional autoregressive models that synthesize points sequentially, our method predicts the Level-of-Detail (LoD) hierarchy of the Gaussian Splatting field and reconstructs the scene in only $\log n$ steps, where $n$ denotes the number of Gaussians.}
\end{center}%
}]

\blfootnote{\textsuperscript{\dag} Equal contribution.}
\blfootnote{\textsuperscript{*} Corresponding authors.}

\begin{abstract}

Auto-regressive frameworks for next-scale prediction of 2D images have demonstrated strong potential for producing diverse and sophisticated content by progressively refining a coarse input. However, extending this paradigm to 3D object generation remains largely unexplored. In this paper, we introduce auto-regressive Gaussian splatting (ARGS), a framework for making next-scale predictions in parallel for generation according to levels of detail. We propose a Gaussian simplification strategy and reverse the simplification to guide next-scale generation. Benefiting from the use of hierarchical trees, the generation process requires only \(\mathcal{O}(\log n)\) steps, where \(n\) is the number of points. Furthermore, we propose a tree-based transformer to predict the tree structure auto-regressively, allowing leaf nodes to attend to their internal ancestors to enhance structural consistency. Extensive experiments demonstrate that our approach effectively generates multi-scale Gaussian representations with controllable levels of detail, visual fidelity, and a manageable time consumption budget.
\end{abstract}    
\section{Introduction}
\label{sec:intro}

3D content generation has experienced remarkable progress in recent years, driven by advances in differentiable rendering and generative modeling. Techniques in volumetric rendering, such as Gaussian splatting (GS)~\cite{kerbl3Dgaussians}, have significantly enhanced our ability to model geometric structures and photorealistic appearances simultaneously from multi-view image inputs. These advancements have led to a new wave of research focused on high-quality object generation. While promising, challenges persist in achieving high fidelity while maintaining real-time rendering efficiency, as well as in representing 3D content in a structured and compact form to facilitate further processing.

% Reconstructing and synthesizing complex 3D scenes with high fidelity and computational efficiency remains one of the core challenges in modern computer vision and computer graphics. With the growing demand for immersive digital content in virtual and augmented reality, the ability to generate 3D assets that are both realistic and lightweight has become increasingly crucial. 
Fortunately, recent advances in Gaussian splatting, or more broadly, Gaussian-based explicit object and scene representations, have demonstrated remarkable potential in modeling volumetric appearance and geometry. By directly encoding a scene as a collection of anisotropic 3D Gaussians, these approaches enable differentiable, high-quality rendering with impressive speed, circumventing the heavy computational costs associated with dense voxel grids or implicit neural radiance fields. Consequently, Gaussian splatting has rapidly emerged as a promising paradigm for efficient 3D reconstruction and novel view synthesis.

In contrast to traditional mesh-based representations, Gaussian splatting offers several unique advantages that make it particularly suitable for modern differentiable rendering pipelines. Classical meshes require explicit connectivity, watertight topology, and carefully maintained surface normals~\cite{kazhdan2006poisson, kato2018renderer, liu2019softras, liu2020general}, which substantially complicate optimization and often obstruct gradient flow when geometry or topology is expected to evolve during training. By comparison, Gaussian primitives are topology-free and continuously parameterized, allowing them to be freely created, removed, or deformed without violating structural constraints or introducing non-differentiabilities.

Moreover, each Gaussian intrinsically stores rich per-primitive appearance attributes—such as color, opacity, scale, and orientation—which can be optimized end-to-end in a fully differentiable manner. In mesh pipelines, adding high-quality view-dependent appearance typically relies on neural textures~\cite{thies2019deferred} or differentiable mesh renderers~\cite{kato2018renderer}, and often requires UV parameterization or carefully constructed surface fields~\cite{park2019deepsdf}. These additional components introduce complexity and can limit flexibility. As a result, Gaussian-based representations provide both a simpler optimization landscape and greater expressiveness in modeling fine-grained, view-dependent appearance, making them a compelling alternative to classical mesh-based approaches.

Despite their impressive efficiency and rendering quality, earlier Gaussian-splatting-based representations still suffer from several fundamental limitations. First, vanilla 3D Gaussian Splatting~\cite{kerbl3Dgaussians} typically requires tens or even hundreds of thousands of Gaussians to accurately model fine geometric structures, resulting in large memory footprints and reduced scalability for complex outdoor or large-scale scenes. Subsequent works attempt to address this through compression or sparse optimization~\cite{fang2023splatfields}, yet the underlying over-parameterization issue remains prominent. 

% Additionally, the optimized base generation is highly sensitive to the quality of the SfM initialization. Noisy camera poses or imperfect point clouds often lead to unstable Gaussian placement, floating primitives, or duplicated structures~\cite{zwicker2002splatting}. Because each Gaussian is optimized independently without explicit structural priors, previous methods struggle to preserve sharp edges, planar surfaces, or topology-consistent geometry, often producing over-smoothed or inconsistent density distributions~\cite{du2023gaussiansurfacereg}.

Furthermore, generative or controllable modeling remains challenging: most methods focus on direct fitting from posed images and do not support hierarchical, level-of-detail, or global-aware operations~\cite{wu2024gaussiandreamer, tang2024splatfacto}. These limitations highlight the need for structured, scalable, and generative formulations of Gaussian representations that move beyond pure per-primitive optimization.

To overcome these challenges, we propose a reversible progressive Gaussian reduction framework that facilitates both top-down simplification and bottom-up synthesis of Gaussian-based scene representations. Specifically, our approach removes Gaussian primitives one by one in a carefully designed, fidelity-preserving, and reversible manner. Each elimination step minimizes perceptual and structural loss, thereby maintaining high reconstruction quality even under aggressive reduction. This stepwise elimination process naturally induces a tree-structured hierarchical representation of the 3D scene, where each node corresponds to a Gaussian, and its descendants represent finer details revealed in subsequent levels. Such a hierarchical structure supports efficient multi-level access, adaptive rendering, and controllable scene complexity, thereby providing a scalable pathway for resource-limited 3D processing.

Inspired by the biological intuition of hierarchical cell division, we leverage the constructed Gaussian hierarchy to develop a next-scale Gaussian generation model that synthesizes finer-level Gaussian configurations conditioned on their coarser ancestors. This autoregressive generative process enables the progressive reconstruction of 3D scenes from low to high resolutions, in a manner reminiscent of multi-scale generation mechanisms used in diffusion and autoregressive image models. By tightly coupling reversible Gaussian reduction with hierarchical generative expansion, our framework unifies compression, reconstruction, and synthesis within a single Gaussian-based representation. This integration offers a principled, interpretable, and scalable foundation for efficient 3D scene modeling.

\noindent Our main contributions are summarized as follows.
\begin{enumerate}
    \item We propose a reversible framework for progressive Gaussian simplification and generation that removes one Gaussian point at a time through a fidelity-preserving elimination process, enabling explicit control over reconstruction quality and representation density.
    \item This progressive reduction process induces a tree-structured hierarchical representation whose complexity grows logarithmically with the number of reduction steps, significantly reducing generation complexity and facilitating multi-scale access and adaptive refinement.
    \item We develop a next-scale Gaussian generation method by leveraging the hierarchical data produced by the reduction process. Our generation can synthesize finer-level Gaussian configurations from coarser representations, enabling multi-resolution reconstruction and hierarchical 3D scene synthesis within a unified framework.
\end{enumerate}

\section{Related Works}
\label{sec:related}
\subsection{Level of Detail and Hierarchical 3D Representations}

Level-of-Detail (LoD) modeling has long been an essential concept in computer graphics and 3D vision, aiming to represent complex scenes with varying geometric and appearance fidelity depending on viewing conditions or computational budgets. Early works in mesh-based LoD \cite{Schroeder1992, Xia1996, Hoppe1996, Hoppe1997, Garland1997} introduced progressive mesh simplification and reconstruction frameworks, allowing transitions between coarse and fine representations. In neural 3D representations, hierarchical structures have been incorporated to improve scalability and rendering efficiency. For instance, Mip-NeRF \cite{barron2022mipnerf360} adopts anti-aliasing and multi-resolution sampling to achieve smooth level transitions in volumetric rendering, while PlenOctrees \cite{yu2021plenoctrees} and Instant-NGP \cite{mueller2022instant} exploit spatial hierarchies such as octrees or hash grids for efficient real-time rendering. Recently, Gaussian Splatting-based methods have also embraced hierarchical or LoD-inspired strategies: Coarse-to-fine densification \cite{kerbl3Dgaussians} adaptively refines Gaussians based on visibility and error metrics, enabling real-time scalability, and GaussianOctree \cite{10670746} introduces a hierarchical tree structure for compact and adaptive Gaussian management. These advances collectively demonstrate that hierarchical LoD principles not only enhance rendering efficiency but also provide a structured pathway for scalable Gaussian generation and progressive reconstruction.

\subsection{Auto-Regressive Model for Next Scale Generation}

Auto-regressive (AR) modeling is a fundamental paradigm that sequentially predicts future elements of a signal or structure conditioned on its past context. In the domain of 3D generative modeling, AR strategies have been increasingly adopted to progressively refine or upscale geometry and appearance across multiple scales, enabling structured and controllable synthesis. VAR \cite{VAR} integrates a VQ-VAE with a Transformer-based autoregressive decoder to model visual information in a discrete latent space, achieving hierarchical image generation. PointNSP~\cite{meng2025pointnspautoregressive3dpoint} extends this paradigm to point clouds, proposing an autoregressive framework that predicts finer-scale point distributions conditioned on coarser inputs, thereby enabling coarse-to-fine 3D shape generation. Similarly, ARMesh~\cite{lei2025armesh} and VertexRegen~\cite{Zhang_2025_ICCV_VertexRegen} explore autoregressive mesh generation by leveraging progressive simplification and its inverse process, providing a bidirectional pathway for structured 3D object and scene synthesis. 
SAR3D~\cite{chen2024sar3d} introduces a multi-scale 3D VQ-VAE that encodes shapes into hierarchical triplane tokens, and performs next-scale autoregressive prediction directly in this structured 3D latent space.

\subsection{Gaussian Splatting Generation}
Unlike NeRFs or SDFs, 3D Gaussian Splatting (3DGS) offers a representation that is much more friendly to rendering, and it carries richer color information than meshes. This makes it a highly promising direction for future 3D development. A growing line of work explores generation, dynamics, and editing of Gaussian Splatting (GS) representations. We review several representative methods as follows:
DreamGaussian \cite{yi2023gaussiandreamer} focuses on fast text or image-conditioned 3D asset generation using progressive Gaussian densification and differentiable rasterization. LGM \cite{tang2024lgm} predicts multi-view Gaussian features via an asymmetric U-Net and fuses them to produce high-resolution GS. Atlas Gaussians \cite{yang2025atlas} represents 3D shapes as multiple local atlas patches, each decoded into a dense set of 3D Gaussians. L3DG \cite{roessle2024l3dg} encodes Gaussian point sets into a VQ-VAE latent space, enabling diffusion over discrete latent tokens rather than raw 3D parameters. DiffGS \cite{DiffGS} reformulates discrete Gaussian sets as continuous functions and applies diffusion in this functional domain.

\section{Method}
\label{sec:method}

\subsection{Preliminary}

Following 3D Gaussian Splatting (3DGS)~\cite{kerbl3Dgaussians}, we define the Gaussian basis function as
\begin{equation}
\small
\label{equ:gs_define}
\begin{aligned}
\mathcal{G}(\mathbf x; o, \mathbf u, \mathbf S, \mathbf R)
&= o \exp\left({ - \tfrac{1}{2} (\mathbf x - \mathbf u)^\top \bm\Sigma^{-1} (\mathbf x - \mathbf u) }\right),
\\
\bm \Sigma &= \mathbf R \mathbf S \mathbf S^{\top} \mathbf R^{\top}, \ \bm \Sigma \succ 0,
\end{aligned}
\end{equation}
where \( \mathbf x \in \mathbb{R}^3 \) denotes a point in 3D space, \( \mathbf u \in \mathbb{R}^3 \) is the Gaussian center,  \( o \in \mathbb{R} \) is its opacity, and \( \mathbf S, \mathbf R \in \mathbb{R}^{3\times 3} \) represent the scaling and rotation of the Gaussian ellipsoid, respectively.
% \xifeng{explain what do these variables mean to make the context self-contained.}.

From the integral form of Eq.~\ref{equ:gs_define}, we derive the closed-form expressions for the zeroth-, first-, and second-order moments of the Gaussian, as shown by Eq.~\ref{equ:moments}. 

\begin{equation}
\label{equ:moments}
\begin{aligned}
m_0 &= o (2\pi)^{3/2} \det(\bm \Sigma)^{1/2}, \\
\mathbf m_1 &= 
% \int_{\mathbb{R}^n} \mathbf x , \mathcal G(\mathbf x) , d\mathbf x 
 m_0 \mathbf u
 .
%  , \\
% \mathbf M_2 &= 
% % \int_{\mathbb{R}^n} \mathbf x \mathbf x^\top \mathcal G(\mathbf x) , d\mathbf x 
%  m_0 \big(\mathbf u \mathbf u^\top + \bm \Sigma\big).
\end{aligned}
\end{equation}

The zeroth-order moment $m_0$ corresponds to the total mass, the first-order moment $\mathbf{m}_1$ indicates that the Gaussian mean coincides with its center $\mathbf{u}$.
% , and the second-order moment $\mathbf{M}_2$ encapsulates both the spatial mean and the anisotropic covariance structure.

\subsection{Simplification Principle}

\begin{figure}
    \centering
    \includegraphics[width=1\linewidth]{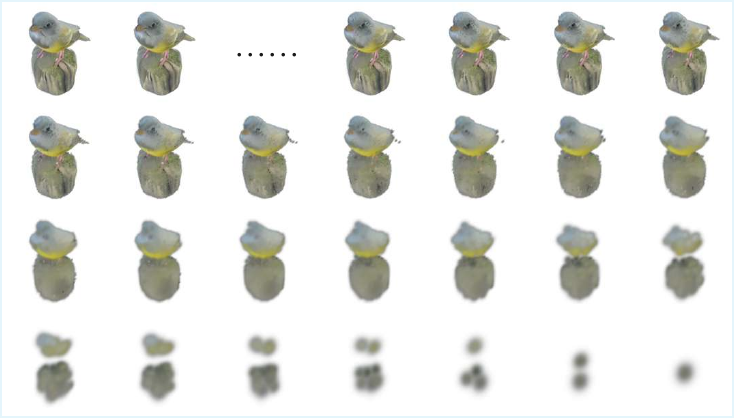}
    \caption{\textbf{A sample of our simplification process.} From left to right and top to bottom, we visualize how a Gaussian Splatting object is progressively downsampled to a single point.}
    \label{fig:tree}
\end{figure}

To determine the merge sequence, we require a principled criterion for selecting which Gaussian to merge. Since Gaussian Splatting models are optimized for visibility and their metrics are primarily visibility-based, we aim to identify the Gaussian that contributes the least to rendering quality. Ideally, this would involve evaluating each Gaussian’s visibility across multiple viewpoints; however, such an approach is computationally expensive. To enable a faster merging process, we seek a rendering-free approximation that remains efficient and effective.

We make the following assumptions:

\begin{enumerate}
    \item The distance between the object and the viewpoints is approximately constant. Therefore, a smaller Gaussian, having a smaller projected area, contributes less to the final rendering.
    \item Merging the least contributing Gaussian with its nearest neighbor minimally affects overall rendering quality.
\end{enumerate}

% Based on these assumptions, we sort all Gaussians by the determinant of their covariance matrices. 
Motivated by these assumptions, we sort all Gaussians according to the determinant of their covariance matrices, where a smaller determinant corresponds to a more compact Gaussian with lower expected visual impact.
At each iteration, the Gaussian with the smallest covariance determinant is merged with its nearest neighbor, forming a new Gaussian component that preserves rendering fidelity while accelerating the merging process. We employ a heap-based priority queue to iteratively merge Gaussian components, where the selection criterion depends on both the determinant of the covariance.
% \xifeng{iteratively merge two Gaussian components, where the selection criterion depends on the determinant of both covariances?}

For each Gaussian with the minimum covariance determinant, we determine its neighbor using a weighted distance metric. Specifically, we compute distances based on the weighted Gaussian properties and select the Gaussian whose distance is closest as its merging partner.

Given two Gaussian components parameterized by \(\mathbf{u}_1, o_1, \mathbf{f}_1, \bm{\Sigma}_1\) and \(\mathbf{u}_2, o_2, \mathbf{f}_2, \bm{\Sigma}_2\), 
where \(\mathbf{f}_i\) denotes the Spherical Harmonics feature, 
% \xifeng{The meaning of these parameters should be stated earlier in the previous subsection}, 
we define a merge operator \(M\) to produce a merged Gaussian as:

\begin{equation}
    [\mathbf{u}_3, o_3, \mathbf{f}_3, \bm{\Sigma}_3]=M(\mathbf{u}_1, o_1, \mathbf{f}_1, \bm{\Sigma}_1, \mathbf{u}_2, o_2, \mathbf{f}_2, \bm{\Sigma}_2)
\end{equation}

While various Gaussian merging strategies exist, we adopt a moment-based weighted formulation for simplicity and efficiency.
% The zeroth- and first-order moments of a Gaussian are expressed as

% \begin{equation}
% m_0 = o (2\pi)^{3/2} |\mathbf{\Sigma}|^{1/2}, \quad
% \mathbf{m}_1 = m_0 \, \mathbf{u}.
% \end{equation}

To account for the overlap between two Gaussians, we introduce an auxiliary (cross) Gaussian defined as:

\begin{equation}
\begin{aligned}
o_c &=o_1o_2,\\
\mathbf{\Sigma}_c &= \big( \mathbf{\Sigma}_1^{-1} + \mathbf{\Sigma}_2^{-1} \big)^{-1}, \\
\mathbf{u}_c &= \mathbf{\Sigma}_c \big( \mathbf{\Sigma}_1^{-1}\mathbf{u}_1 + \mathbf{\Sigma}_2^{-1}\mathbf{u}_2 \big)    
\end{aligned}
\end{equation}
.
% , \\
% o_c &= o_1 o_2

The merged opacity\cite{mildenhall2020nerf} and moments are then given by
\begin{equation}
\begin{aligned}
o_3 &= o_1 + o_2 - o_c, \\
m_{0,3} &= m_{0,1} + m_{0,2} - m_{0,c}, \\
\mathbf{m}_{1,3} &= \mathbf{m}_{1,1} + \mathbf{m}_{1,2} - \mathbf{m}_{1,c},
\end{aligned}    
\end{equation}
from which the new mean is obtained as \(\mathbf{u}_3 = \mathbf{m}_1 / m_0\). The notation $m_{i,j}$ denotes the $i$-th moment of the $j$-th Gaussian point (where the third, $j=3$, represents the merged point), and $m_{i,c}$ denotes the $i$-th moment of the auxiliary (cross) Gaussian.

The merged covariance is computed via a moment-weighted average:
\begin{equation}
\mathbf{\Sigma}_3
= \frac{ m_{0,1}\mathbf{\Sigma}_1 + m_{0,2} \mathbf{\Sigma}_2}{m_0}
\end{equation}

If each Gaussian carries a feature vector \(\mathbf{f}_i\), we perform feature fusion using moment-weighted averaging:
\begin{equation}
\mathbf{f}_3 = 
\frac{m_{0,1}\mathbf{f}_1 + m_{0,2}\mathbf{f}_2}{m_{0,1}+m_{0,2}}.
\end{equation}
% The final merged Gaussian is represented as \((\mathbf{u}_3, o_3, \mathbf{f}_3, \mathbf{\Sigma}_3)\).

\subsection{Hierarchical Spatial Structure}
\label{sec:hier_spatial_structure}

\begin{figure}
    \centering
    \includegraphics[width=1\linewidth]{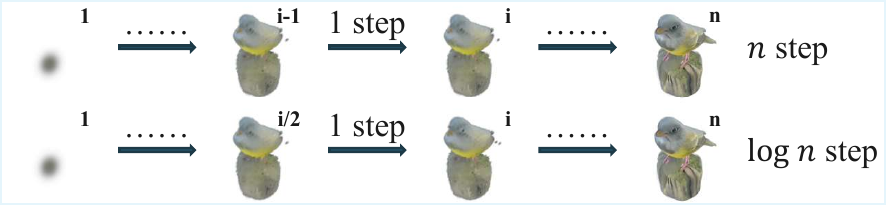}
    \caption{\textbf{Comparison of autoregressive generation steps.} Numbers on the top right indicate the number of Gaussian points.
    \textbf{Top:} Vanilla autoregressive (AR) models predict one token at a time; therefore, generating \(n\) points requires \(n-1\) sequential steps. 
    \textbf{Bottom:} Our hierarchical AR model predicts the next level in a spatial hierarchy, where each level expansion generates multiple points in parallel. This hierarchical formulation reduces the generation complexity from linear to logarithmic, requiring only \(\log n\) steps.
    }
    \label{fig:ar_steps}
\end{figure}

\begin{figure*}
    \centering
    \includegraphics[width=0.8\linewidth]{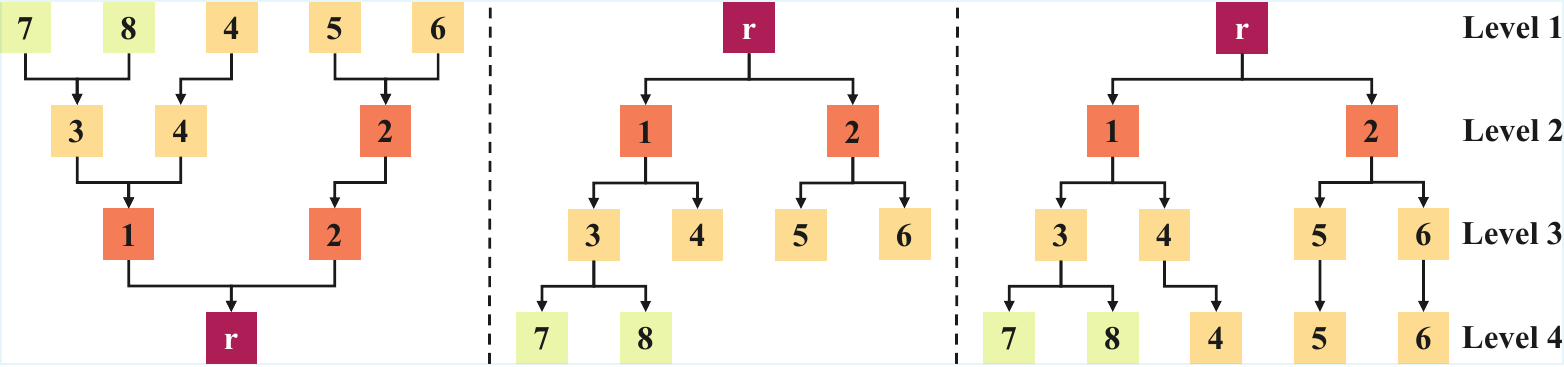}
    \caption{
    \textbf{Hierarchical Spatial Structure.} 
    \textbf{Left:} The simplification process merges node pairs iteratively to form higher-level nodes; independent pairs can be merged in parallel. 
    \textbf{Middle:} The binary tree view reverses the simplification process, reconstructing the hierarchy from the root and removing duplicate nodes to obtain level-wise data. 
    \textbf{Right:} The hierarchical spatial tree representation stores only the leaf nodes at each level, discarding internal nodes that have already been split.
    }
    \label{fig:hier_spatial_structure}
\end{figure*}

Directly fitting the original merge sequence entails a linear computational complexity of \( \mathcal{O}(n) \), where $n$ denotes the number of points. This is because each autoregressive step generates precisely one additional point, resulting in a total of \(n-1\) sequential steps to complete the process, as illustrated in Fig.~\ref{fig:ar_steps} (top).
To improve scalability, we propose a hierarchical generation graph that enforces a multi-level spatial structure, enabling the model to predict multiple Gaussian splits concurrently. This formulation substantially reduces the number of auto-regressive steps required for generation.
The hierarchy is initialized from the root node representing the final merged Gaussian and is recursively expanded by locating each node’s two children and replacing the parent with its descendants until the full structure is recovered, as shown in Fig.~\ref{fig:hier_spatial_structure} (right).

Formally, given an input Gaussian set 
\(\mathcal{P} = \{p_i\}_{i=1}^n\),
we define the merge sequence 
\(\mathcal{M} = \text{Simplify}(\mathcal{P})\),
which is then reversed for root-to-leaf traversal. 
For each merge operation \(m \in \mathcal{M}\), we construct a parent-to-child mapping:
\begin{equation}
\varsigma[m_{\text{parent}}] = [\, m_{\text{child1}},\ m_{\text{child2}} \,].
\end{equation}
This means that a Gaussian $m_{\text{parent}}$ can be split into two Gaussians, $m_{\text{child1}}$ and $m_{\text{child2}}$, as its children.
Starting from the root set \(\mathcal{N}_0 = \{r\}\), where  is the final Gaussian after all the merges, each subsequent level is recursively obtained as
\begin{equation}
\mathcal{N}_{l+1} =
\bigcup_{n \in \mathcal{N}_l}
\begin{cases}
\varsigma[n], & \text{if } n \text{ can split},\\
n, & \text{otherwise.}
\end{cases}
\end{equation}

The expansion continues until all nodes become indivisible, yielding a hierarchical representation
\(\{ \mathcal{N}_0, \mathcal{N}_1, \dots, \mathcal{N}_L \}\),
where \(L\) denotes the total number of levels. 
This hierarchical formulation transforms the original linear complexity \( \mathcal{O}(n) \) into a logarithmic one \( \mathcal{O}(\log n) \), substantially accelerating Gaussian generation while maintaining structural coherence, as illustrated in Fig.~\ref{fig:ar_steps} (bottom).

\subsection{Tree-based Auto-Regressive Generation}
\label{sec:auto_regressive_generation}

Building upon the hierarchical structure described in Sec.~\ref{sec:hier_spatial_structure}, we model the next-scale prediction as an auto-regressive process, where the probability of the current Gaussian \( \mathcal{N}_t \) depends only on its preceding elements \( (\mathcal{N}_1, \mathcal{N}_2, \ldots, \mathcal{N}_{t-1}) \). This unidirectional dependency assumption enables the factorization of the joint distribution as

\begin{equation}
p(\mathcal{N}_1, \ldots, \mathcal{N}_T)
= \prod_{t=1}^{T} p(\mathcal{N}_t \mid \mathcal{N}_1, \ldots,\mathcal{N}_{t-1}).
\end{equation}

The auto-regressive model \( p_\theta \) is trained to maximize the likelihood of observing the next Gaussian conditioned on its predecessors, i.e.,
\begin{equation}
\max_\theta \sum_{t=1}^{T} \log p_\theta(\mathcal{N}_t \mid \mathcal{N}_1, \ldots, \mathcal{N}_{t-1}).
\end{equation}
This formulation corresponds to the \textbf{next-scale prediction} objective, where each Gaussian token is sequentially predicted from coarser to finer levels within the hierarchy.

During inference, the model generates new Gaussian sequences in a progressive manner. Starting from the root node, each subsequent scale is synthesized by sampling from the learned conditional distribution 
\begin{equation}
p_\theta(\mathcal{N}_t \mid \mathcal{N}_1, \ldots, \mathcal{N}_{t-1}) 
\end{equation}

This process continues until all levels of the hierarchy are expanded, yielding a complete Gaussian representation that preserves both structural consistency and spatial detail.

\begin{figure}
    \centering
    \includegraphics[width=1.0\linewidth]{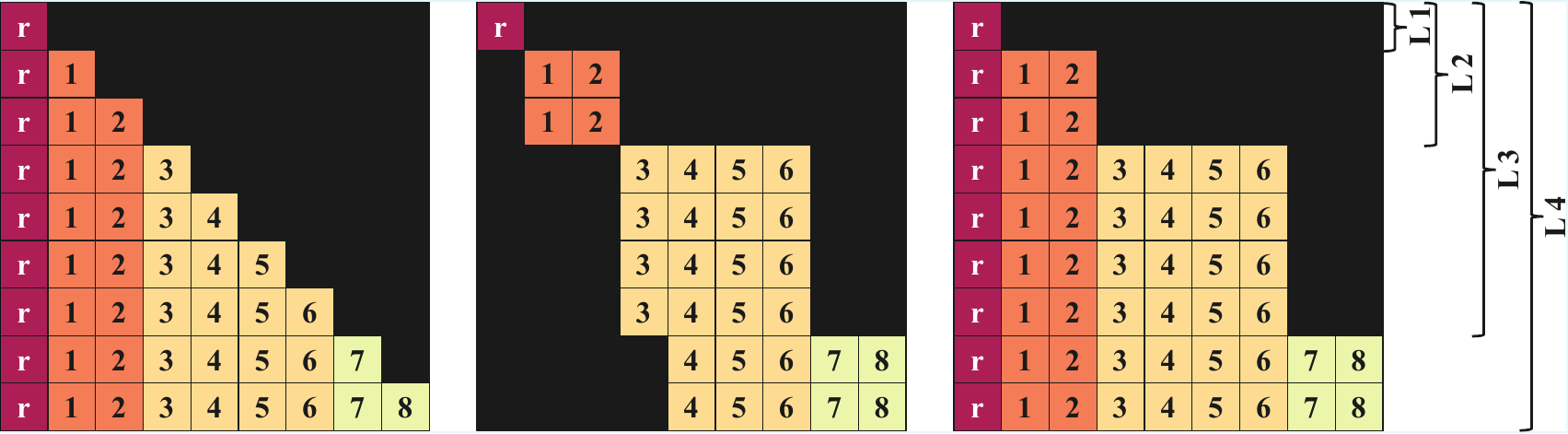}
    \caption{
    \textbf{Attention Mask.} 
    \textbf{Left}: The causal attention mask. It depends on a sorted sequence and generates tokens sequentially, requiring $n$ steps in total. 
    \textbf{Middle}: The level-wise attention mask. It attends only to the leaf nodes within each level, reducing the generation complexity to $\log n$ steps. 
    \textbf{Right}: The tree-based attention mask. It extends level-wise attention by also considering internal nodes from previous levels, allowing token generation within $\log n$ steps.
    The rightmost panel illustrates how many tokens are required to decode each level of detail.
    }
    \label{fig:attention_mask}
\end{figure}

In contrast to the conventional causal attention mask, as illustrated in Fig.~\ref{fig:attention_mask} (left), which enforces strictly sequential token generation and incurs a computational cost of $O(n)$, we introduce a more efficient hierarchical attention framework. Specifically, we propose a hierarchical spatial attention mechanism, shown in Fig.~\ref{fig:attention_mask} (middle), that performs level-wise self-attention over a multi-scale representation of the input. Tokens at each hierarchical level are processed in parallel, reducing the generation complexity to $O(\log n)$ steps.

Building upon this design, we further introduce a tree-based attention mechanism, illustrated in Fig.~\ref{fig:attention_mask} (right), which extends hierarchical spatial attention by incorporating internal nodes from previous levels into the attention computation. This design enables the model to capture long-range dependencies without compromising efficiency, achieving autoregressive generation in $O(\log n)$ steps.

\begin{figure}
    \centering
    \includegraphics[width=0.7\linewidth]{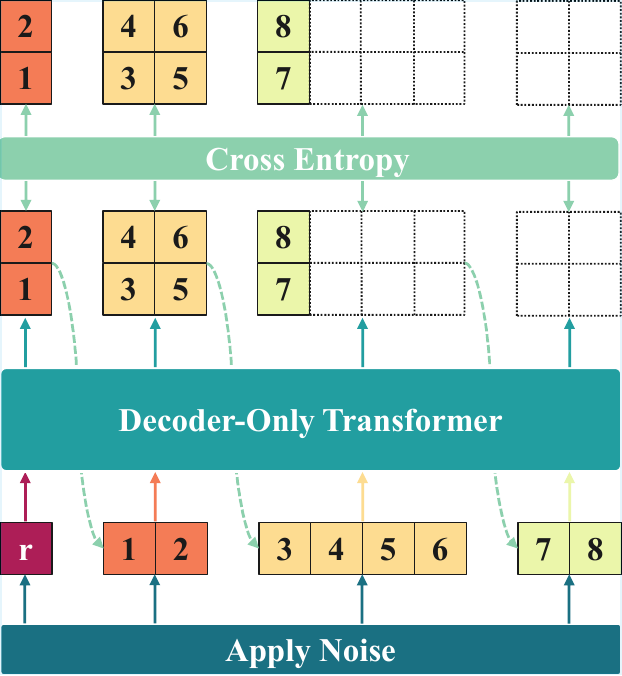}
    \caption{\textbf{Method.} Starting from a root node $\mathbf{r}$, the decoder-only transformer predicts whether each node should be split and generates its corresponding child nodes. The process continues recursively until all nodes become leaf nodes. The model is trained using cross-entropy loss between the predicted and ground-truth nodes.}
    \label{fig:method}
    \vspace{-0.5cm}
\end{figure}

Starting from a root $\mathbf{r}$, 
% \xifeng{capital $R$ to keep it consistent with those shown in figures?}
the decoder-only Transformer recursively predicts the hierarchical structure by expanding each node into its corresponding children. The model first generates the children of $\mathbf{r}$, replaces $\mathbf{r}$ with its descendants, and continues recursively until all nodes become leaf nodes. This auto-regressive procedure naturally models the hierarchical generative process in a sequential manner. The model is optimized using a standard cross-entropy loss between predicted and ground-truth node states, as illustrated in Figure~\ref{fig:method}.

% \subsection{Noise-Augmented Training} 

% We observe that teacher-forcing training introduces a significant train-test discrepancy in autoregressive generation. During training, the Transformer is conditioned only on ground-truth Gaussians at each scale, preventing it from learning to correct its own prediction errors. Consequently, inaccuracies that arise at coarser scales propagate and amplify through subsequent finer scales, ultimately degrading the overall generation quality.

% To address this issue, we propose a noise-augmented training strategy \cite{Infinity} that enhances the model’s robustness to prediction errors. Given a batch of Gaussians, we randomly perturb a subset with a ratio \(0<\alpha<1\), modifying their attributes (\textit{e.g.}, position, opacity, or covariance) while keeping the ground-truth supervision unchanged. This strategy encourages the model to identify and correct corrupted or inconsistent inputs during training, effectively enabling it to recover from its own errors during inference.

\section{Experiments}
\label{sec:exp_and_res}

\subsection{Experimental Setup}

\begin{figure*}[t]
  \centering
  % --- SSIM ---
  \begin{subfigure}[t]{0.32\linewidth}
    \centering
    \includegraphics[width=\linewidth]{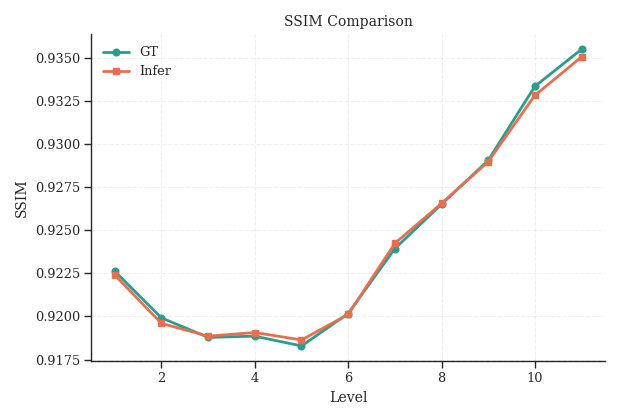}
    \caption{SSIM Comparison}
    \label{fig:ssim}
  \end{subfigure}\hfill
  % --- LPIPS ---
  \begin{subfigure}[t]{0.32\linewidth}
    \centering
    \includegraphics[width=\linewidth]{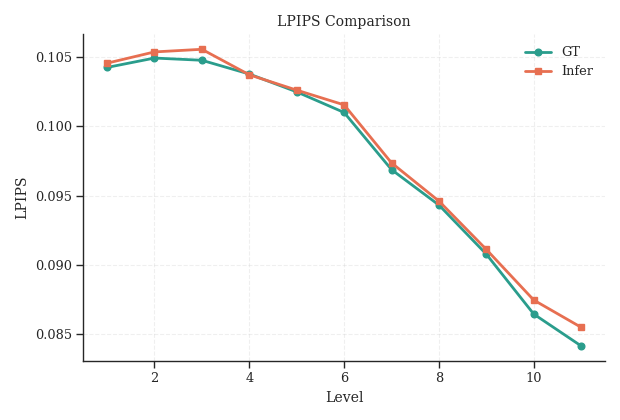}
    \caption{LPIPS Comparison}
    \label{fig:lpips}
  \end{subfigure}\hfill
  % --- PSNR ---
  \begin{subfigure}[t]{0.32\linewidth}
    \centering
    \includegraphics[width=\linewidth]{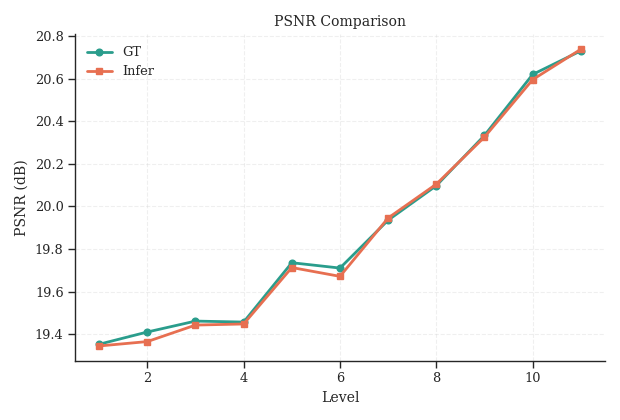}
    \caption{PSNR Comparison}
    \label{fig:psnr}
  \end{subfigure}

  \caption{Level-wise reconstruction quality (GT vs. Infer). 
  Higher is better for SSIM/PSNR; lower is better for LPIPS. 
  Curves start from Level~1.}
  \label{fig:quality_curves}
\end{figure*}

To validate our method from simplification to generation, we conduct a series of experiments. First, we evaluate the simplification method. Second, we test our generation method on selected datasets. Finally, we apply our approach to category-specific large-scale training.

\noindent \textbf{Datasets.} To obtain the merge sequences, we require datasets containing extensive Gaussian Splatting data. ~\cite{ma2025large} propose a large dataset for self-supervised pretraining, which consists of 52,121 ShapeSplat\cite{shapenet2015} objects, 12,309 ModelSplat\cite{wu20153d} objects, and 141,703 Objaverse\cite{objaverse} objects. For training, we select specific categories from ShapeSplat and ModelSplat.

% To enhance data diversity and robustness, we design a Gaussian field augmentation pipeline operating directly on point-based representations. 
% Given a Gaussian field \((\mathbf{x}, o, \mathbf{f}, \mathbf{s}, \mathbf{q})\), the module stochastically perturbs positions, opacities, colors, scales, and rotations.

% Specifically, Gaussian centers are jittered according to their local scales, and color features (in spherical harmonics form) are enhanced using differentiable \texttt{ColorJitter} operations with adjustable brightness, contrast, saturation, and hue. 
% Random rotations, flips, and scaling within a controlled range \([0.8, 1.2]\) are applied to improve geometric variability. 
% Rotations are updated via quaternion multiplication to maintain smoothness and avoid singularities.

% Overall, the augmentation can be formulated as
% \begin{equation}
% (\mathbf{x}', o', \mathbf{f}', \mathbf{s}', \mathbf{q}') 
% = \mathcal{A}(\mathbf{x}, o, \mathbf{f}, \mathbf{s}, \mathbf{q}; p, \theta),
% \end{equation}
% where \(\mathcal{A}\) denotes the stochastic augmentation operator with probability \(p\) and intensity parameters \(\theta\). 
% This process increases geometric and photometric diversity, improving the model’s generalization to varied shapes and lighting conditions.

\noindent \textbf{Metrics.} To evaluate the simplification method, we compare the original Gaussians with their simplified counterparts. We measure rendering quality using PSNR, SSIM, and LPIPS. To assess the quality of sequence fitting, we report both the fitting accuracy of the sequences and the corresponding rendering quality using the same metrics as for the simplification evaluation.
% We additionally report Chamfer-L2 between mesh extractions from rendered depth (uniform marching cubes of the accumulated opacity field) when applicable.

\subsection{Implementation details}

We adopt a Transformer architecture as our backbone.

\noindent \textbf{Network.} Our model consists of 24 Transformer layers, each processing tokens with an embedding dimension of 1536 and employing 24 attention heads. The output includes a boolean indicating whether a Gaussian is splittable and 14×2 sets of 256-class classification logits.

\noindent \textbf{Quantization.} To formulate the problem as classification, all continuous Gaussian attributes are quantized into discrete codebook indices. Each parameter is discretized into 256 bins, and for the scale parameter, we apply a logarithmic transformation prior to quantization. Since Gaussian scales vary widely and are densely distributed at smaller magnitudes, log-space quantization provides a balanced representation across scales, allowing the model to capture both coarse and fine-scale variations effectively.

\noindent \textbf{Token Representation.} Each quantized attribute is embedded via a learnable lookup table. For each Gaussian token, embeddings of position, opacity, feature, scale, and rotation are concatenated and projected into a unified embedding space. These fused embeddings are then linearly mapped into multiple \emph{query}, \emph{key}, and \emph{value} vectors for self-attention computation.

\noindent \textbf{3D-Aware Positional Encoding.} To enhance 3D spatial reasoning, we employ a three-dimensional Rotary Positional Encoding module. Each 3D coordinate is decomposed into $(x, y, z)$ components, with independent rotary encodings applied across attention head dimensions. This preserves spatial equivariance and enables the attention mechanism to capture local geometric correlations effectively.

\noindent \textbf{Noise-Augmented Training.} We observe that teacher-forcing training introduces a significant train-test discrepancy in autoregressive generation. During training, the Transformer is conditioned only on ground-truth Gaussians at each scale, preventing it from learning to correct its own prediction errors. Consequently, inaccuracies that arise at coarser scales propagate and amplify through subsequent finer scales, ultimately degrading the overall generation quality. 
To address this issue, we propose a noise-augmented training strategy \cite{Infinity} that enhances the model’s robustness to prediction errors. Given a batch of Gaussians, we randomly perturb a subset with a ratio \(0<\alpha<1\), modifying their attributes while keeping the ground-truth supervision unchanged. This strategy encourages the model to identify and correct corrupted or inconsistent inputs during training, effectively enabling it to recover from its own errors during inference.

\subsection{Performance of Simplification}
\label{sec:simplification}

\begin{figure}
    \centering
    \includegraphics[width=1.0\linewidth]{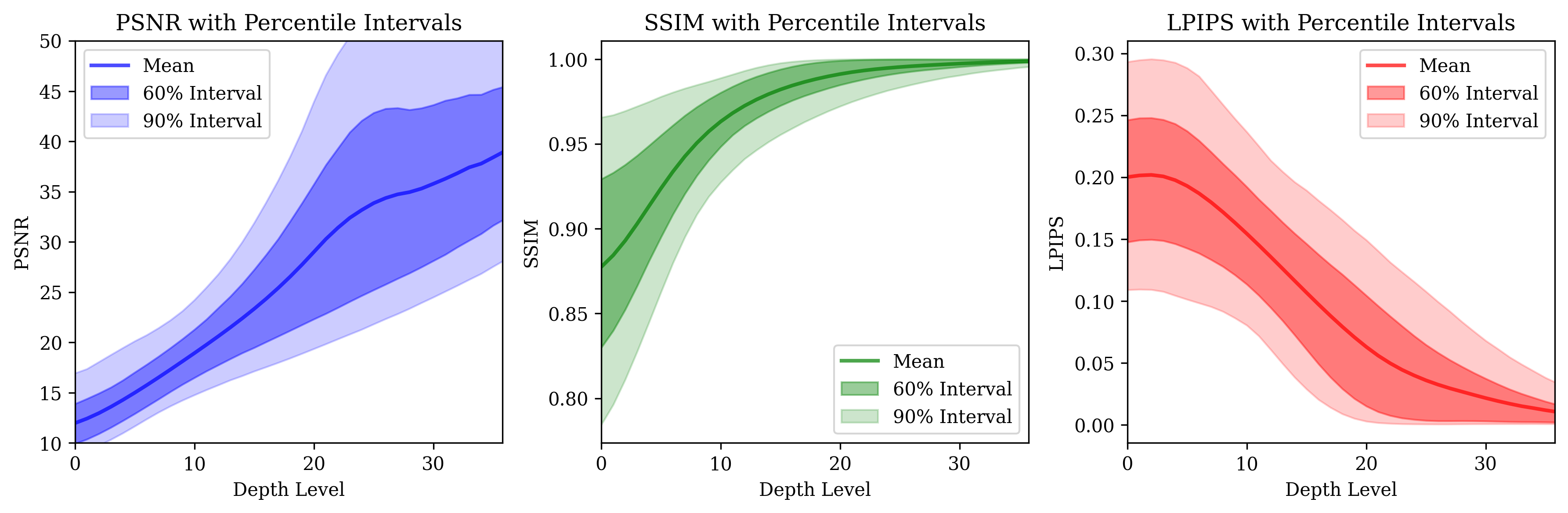}
    \caption{\textbf{Quality of Simplification Method.} We evaluate the quality of the simplified data by comparing it to the ground truth using PSNR, SSIM, and LPIPS. We also report the 60\% and 90\% data distribution to highlight overall trends.}
    \label{fig:simplification}
\end{figure}

To evaluate the proposed simplification method, we compare the simplified objects with their original, unsimplified counterparts. For each simplification level, we render 9,842 \texttt{ModelSplat} objects from 8 viewpoints and compute PSNR, SSIM, and LPIPS against the ground truth. We report both the mean and the 90th-percentile interval for each metric to quantify performance across the dataset, as shown in Fig.~\ref{fig:simplification}.

\subsection{Performance of Single-Object Generation}
\label{sec:single_object}

To assess the fitting capability of our model, we apply it to the \texttt{bird} object and compare the reconstructed results with the ground truth. Quantitative comparisons are presented in Table~\ref{tab:infer_vs_gt}, demonstrating that our model effectively captures the underlying geometry and appearance, confirming its strong ability to fit individual objects.

\begin{table}[t]
\tiny
\centering
\caption{Comparison between inference results and ground-truth reconstruction across hierarchy levels of one single object ``bird".}
\label{tab:infer_vs_gt}
\scalebox{1.0}{
\begin{tabular}{c | r@{\extracolsep{4pt}}r@{\extracolsep{4pt}}r | r@{\extracolsep{4pt}}r@{\extracolsep{4pt}}r | r@{\extracolsep{4pt}}r@{\extracolsep{4pt}}r}
\toprule
\multirow{2}{*}{Level} &
\multicolumn{3}{c|}{Infer} &
\multicolumn{3}{c|}{GT} &
\multicolumn{3}{c}{$\Delta$ (Infer - GT)} \\
% \cmidrule(lr){2-4} \cmidrule(lr){5-7} \cmidrule(lr){8-10}
& PSNR & SSIM & LPIPS
& PSNR & SSIM & LPIPS
& $\Delta$ PSNR & $\Delta$ SSIM & $\Delta$ LPIPS \\
\midrule
1  & 16.37 & 0.9284 & 0.2548 & 16.45 & 0.9291 & 0.2574 & -0.08 & -0.0007 & -0.0026 \\
2  & 18.00 & 0.9228 & 0.3088 & 18.01 & 0.9230 & 0.3111 & -0.01 & -0.0002 & -0.0022 \\
3  & 17.62 & 0.9092 & 0.2705 & 17.60 & 0.9082 & 0.2688 & +0.02 & 0.0010 & +0.0017 \\
4  & 17.99 & 0.9083 & 0.2887 & 18.02 & 0.9080 & 0.2873 & -0.03 & +0.0003 & +0.0014 \\
5  & 18.13 & 0.8900 & 0.2302 & 17.89 & 0.8899 & 0.2300 & +0.24 & +0.0001 & +0.0002 \\
6  & 20.92 & 0.9066 & 0.2034 & 21.20 & 0.9057 & 0.2034 & -0.28 & +0.0009 & 0.0000 \\
7  & 19.85 & 0.9180 & 0.1131 & 19.89 & 0.9183 & 0.1119 & -0.04 & -0.0003 & +0.0012 \\
8  & 21.48 & 0.9244 & 0.0980 & 21.36 & 0.9246 & 0.0925 & +0.12 & -0.0002 & +0.0055 \\
9  & 21.00 & 0.9209 & 0.0896 & 20.91 & 0.9206 & 0.0877 & +0.09 & +0.0003 & +0.0019 \\
10 & 22.56 & 0.9280 & 0.0957 & 22.56 & 0.9278 & 0.0962 & 0.00  & +0.0002 & -0.0005 \\
11 & 22.77 & 0.9229 & 0.1346 & 22.63 & 0.9228 & 0.1348 & +0.14 & +0.0001 & -0.0002 \\
12 & 21.90 & 0.9390 & 0.1367 & 21.82 & 0.9390 & 0.1368 & +0.08 & 0.0000  & -0.0001 \\
13 & 20.76 & 0.9347 & 0.1074 & 20.76 & 0.9348 & 0.1072 & 0.00  & -0.0001 & +0.0002 \\
14 & 22.66 & 0.9503 & 0.1405 & 22.65 & 0.9504 & 0.1409 & +0.01 & -0.0001 & -0.0004 \\
15 & 23.33 & 0.9534 & 0.1410 & 23.22 & 0.9537 & 0.1423 & +0.11 & -0.0003 & -0.0013 \\
\bottomrule
\end{tabular}
}
\end{table}

\subsection{Performance of Single-Class Generation}
\label{sec:single_class}

\begin{figure}
    \centering
    \includegraphics[width=0.9\linewidth]{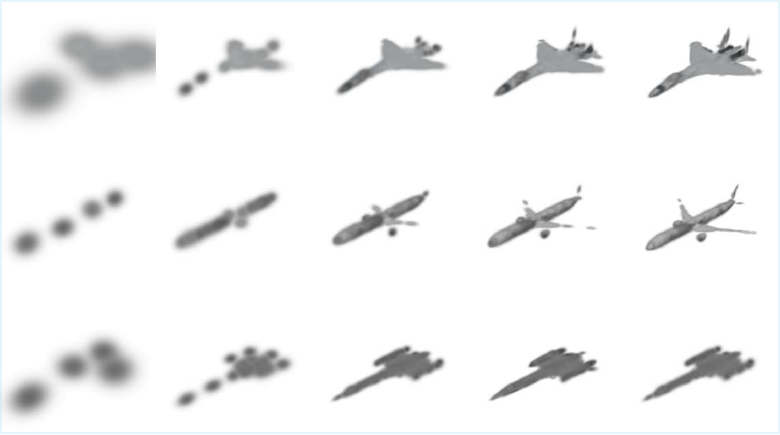}
    \caption{Progressive generation within a single category. Starting from an initial point, the model progressively synthesizes the Gaussian Splatting representation. We visualize intermediate results at steps 2, 4, 7, 9, and 11.}
    \label{fig:progressive_gen_single}
\end{figure}

\begin{figure}[htbp]
    \centering
    \setlength{\tabcolsep}{2pt}
    \renewcommand{\arraystretch}{0}
    \begin{tabular}{cccc}
        \includegraphics[width=0.24\linewidth]{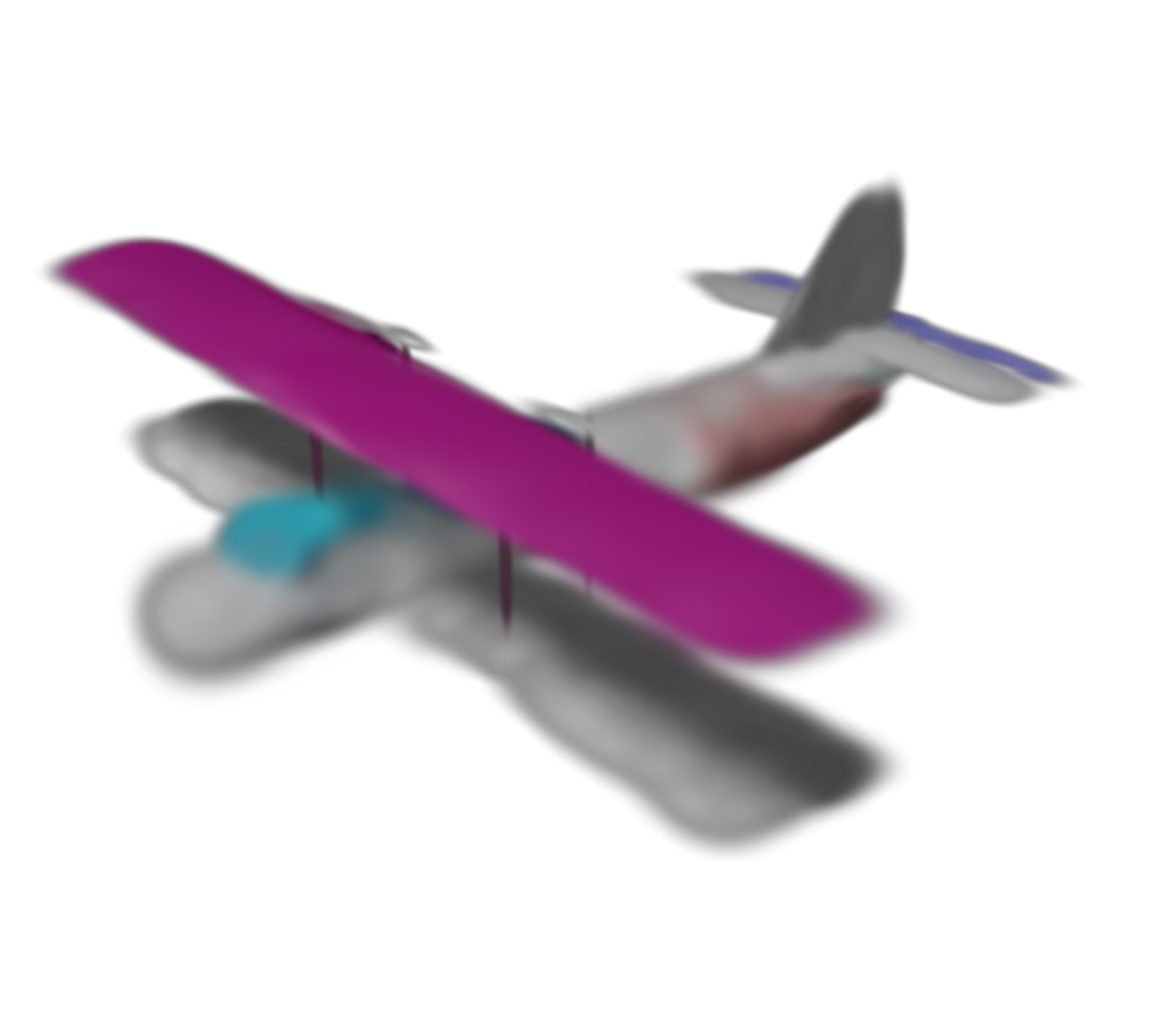} &
        \includegraphics[width=0.24\linewidth]{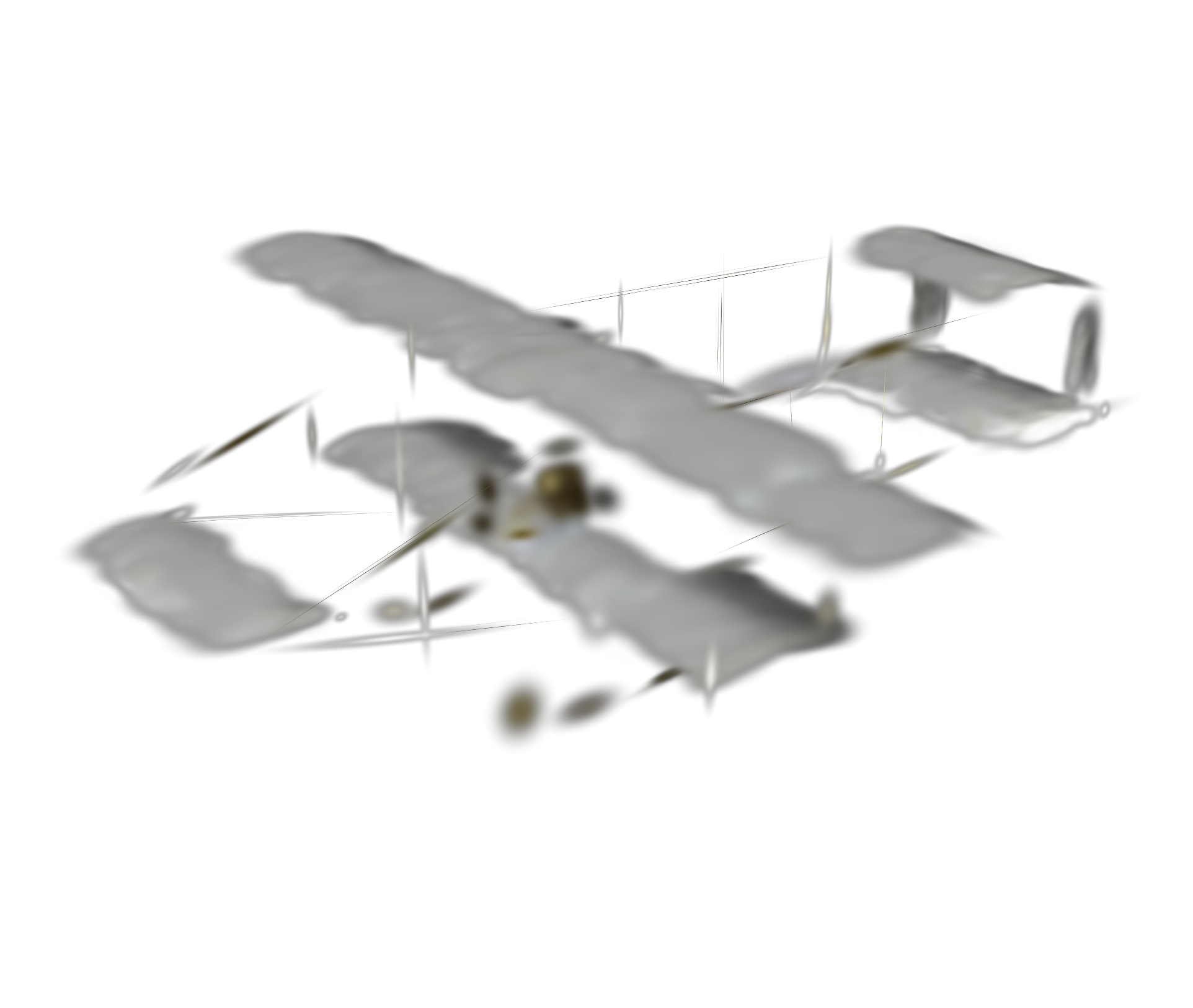} &
        \includegraphics[width=0.24\linewidth]{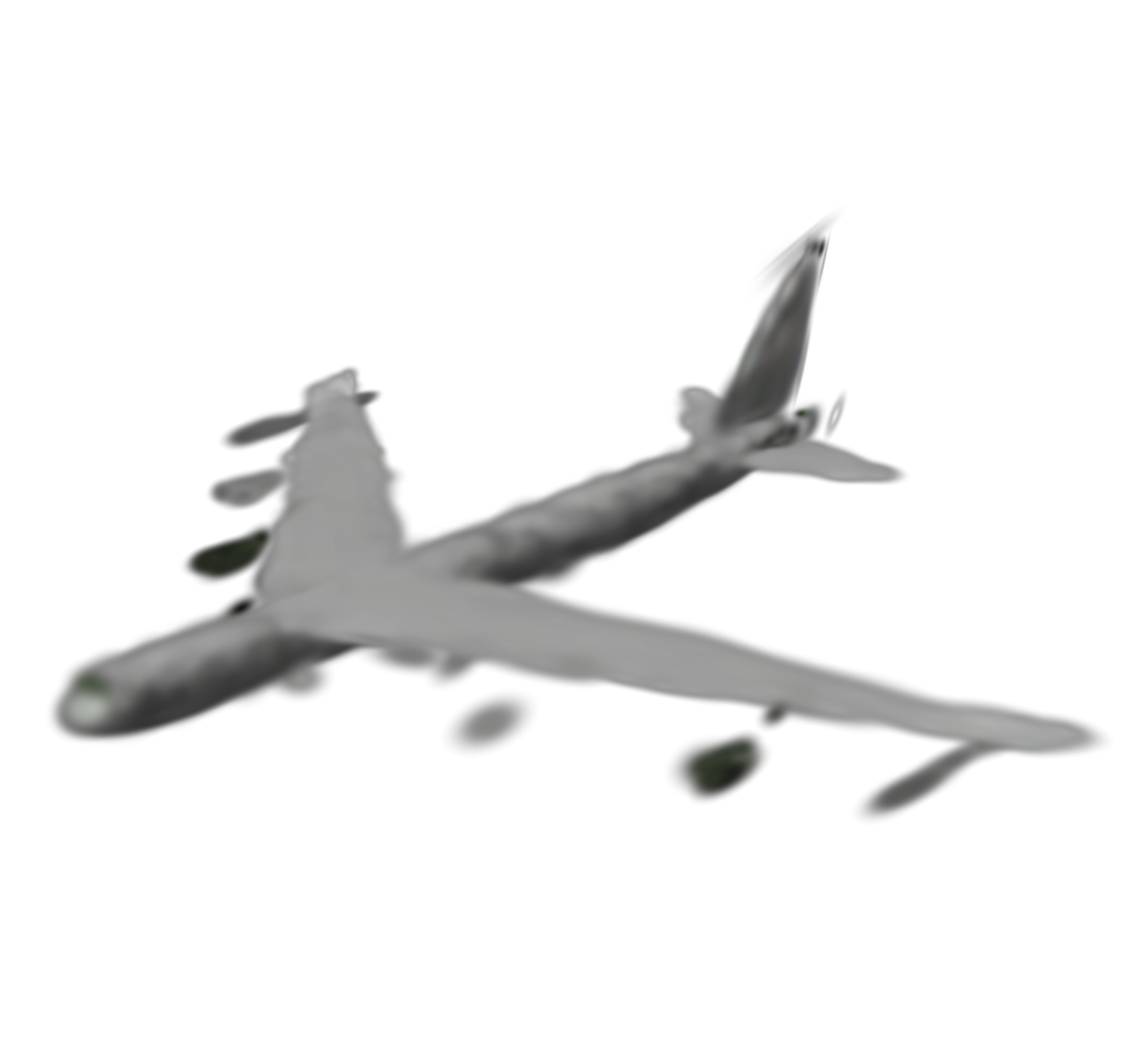} &
        \includegraphics[width=0.24\linewidth]{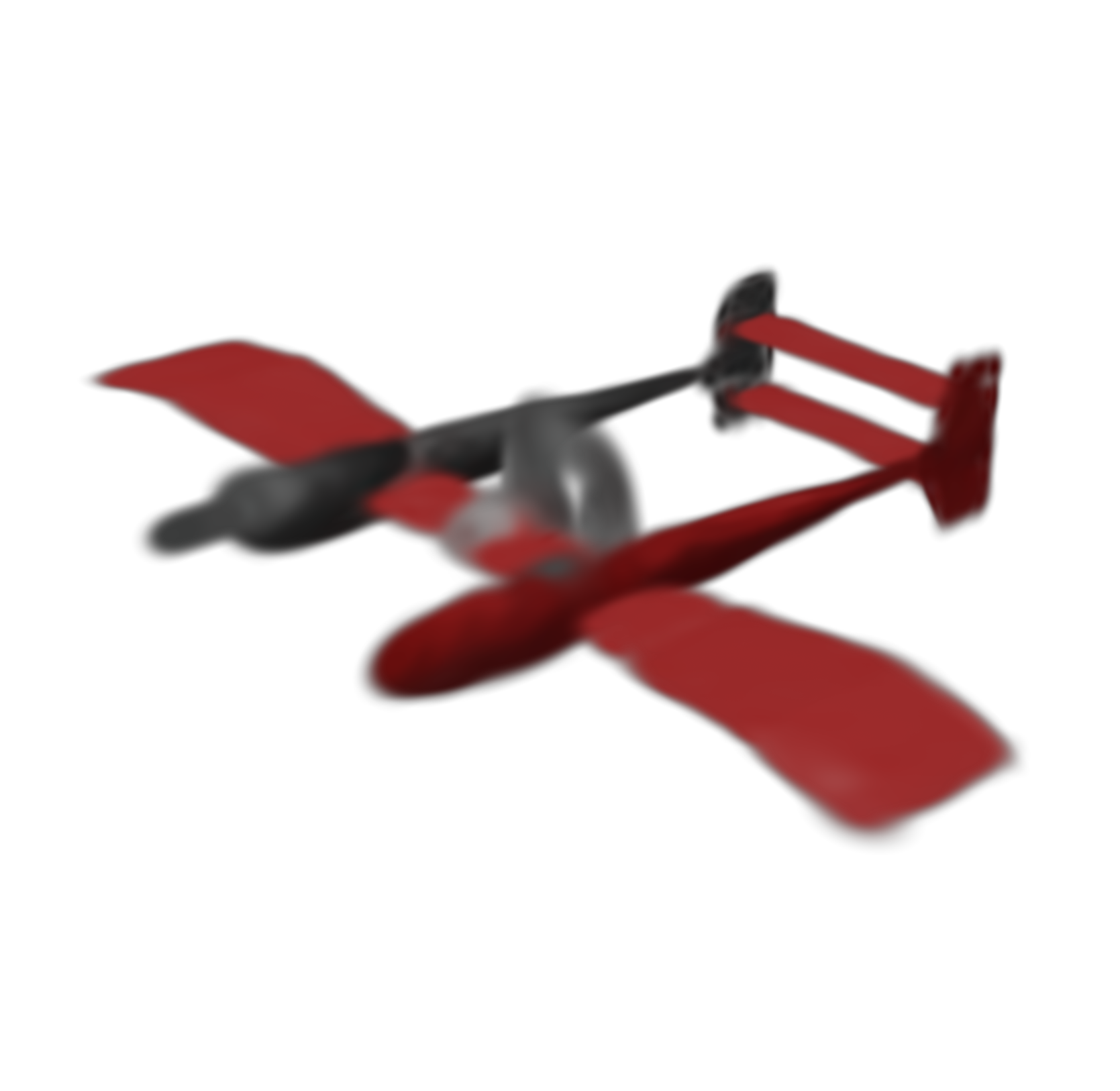} \\
        \includegraphics[width=0.24\linewidth]{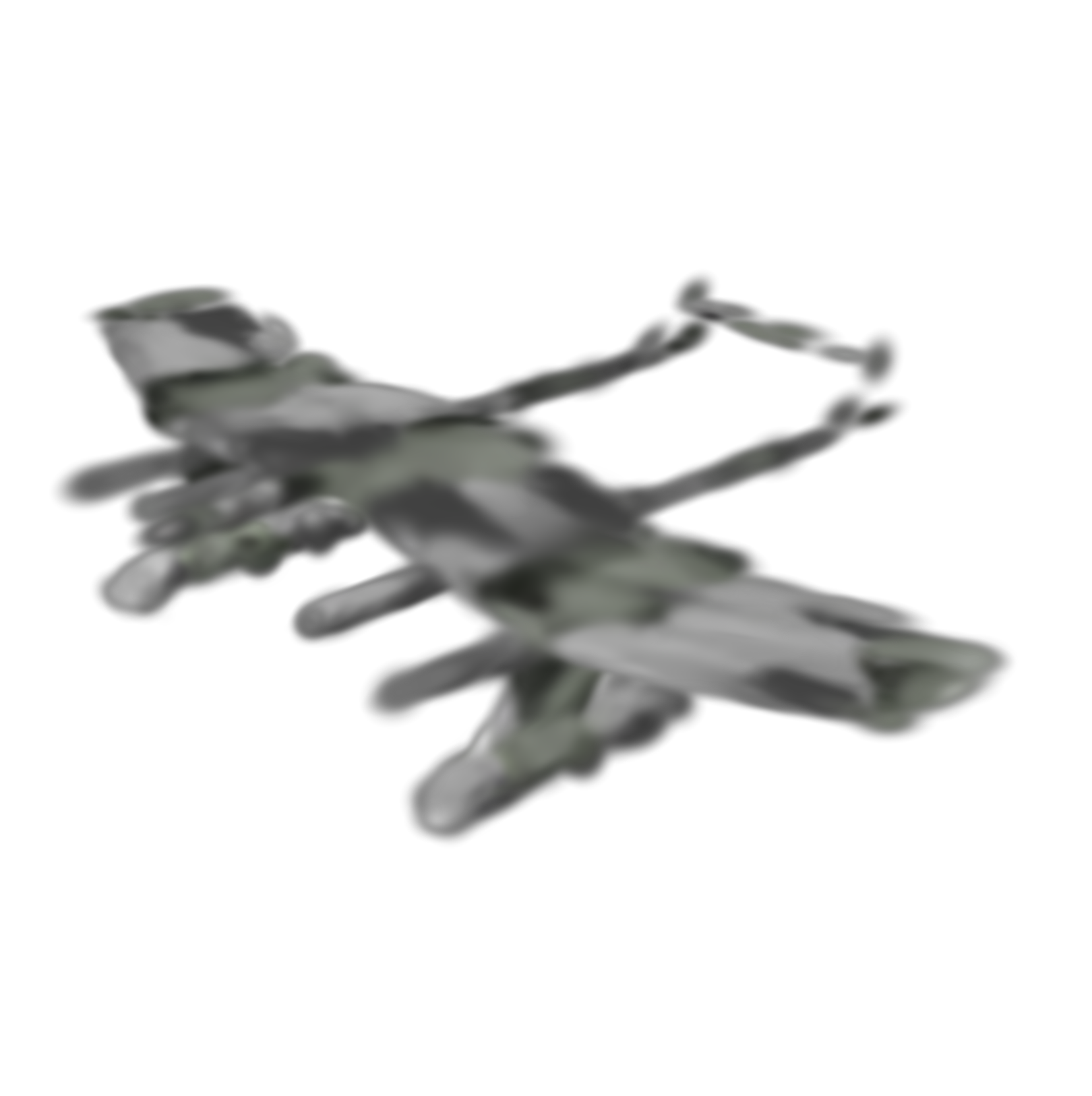} &
        \includegraphics[width=0.24\linewidth]{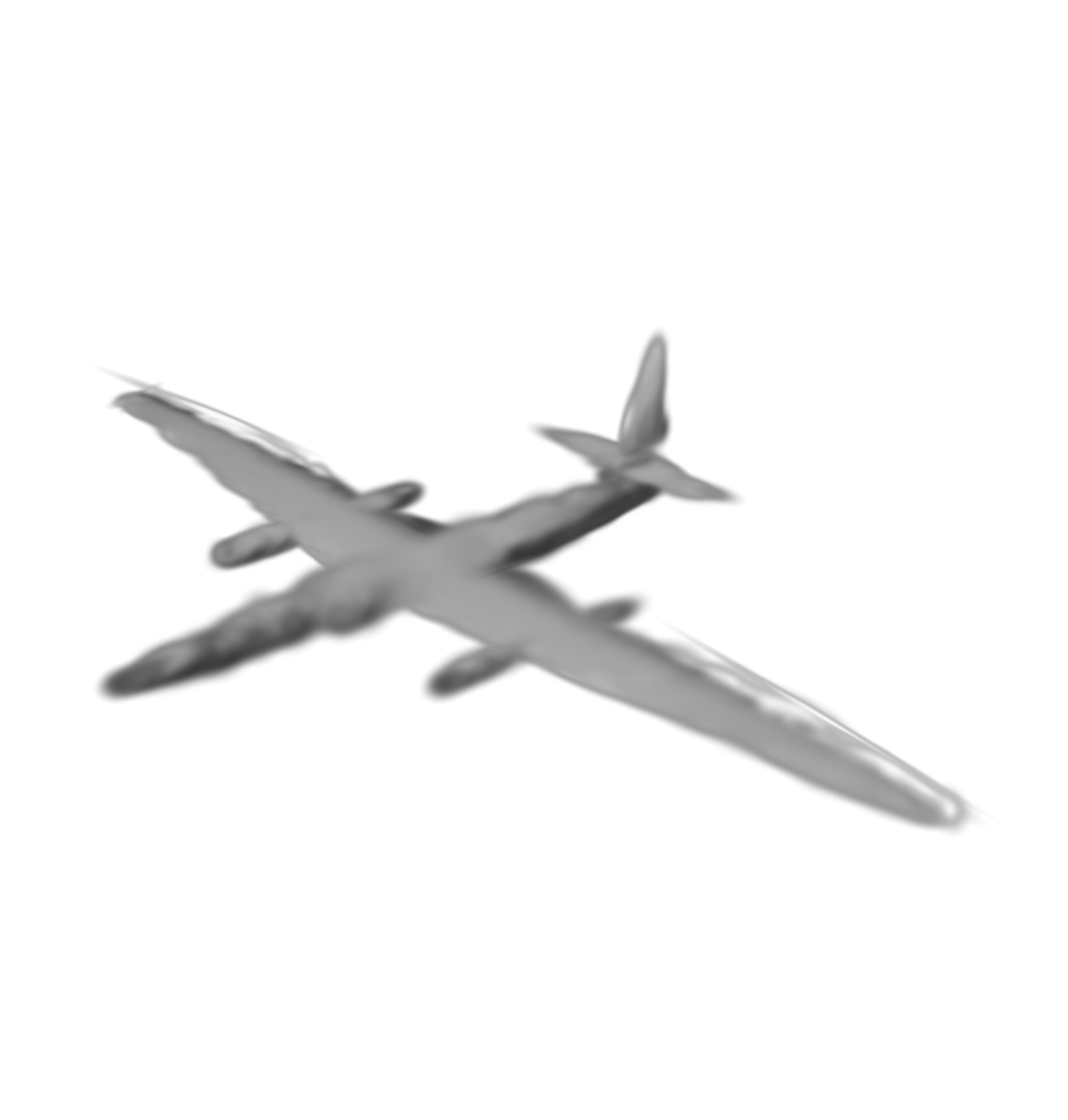} &
        \includegraphics[width=0.25\linewidth]{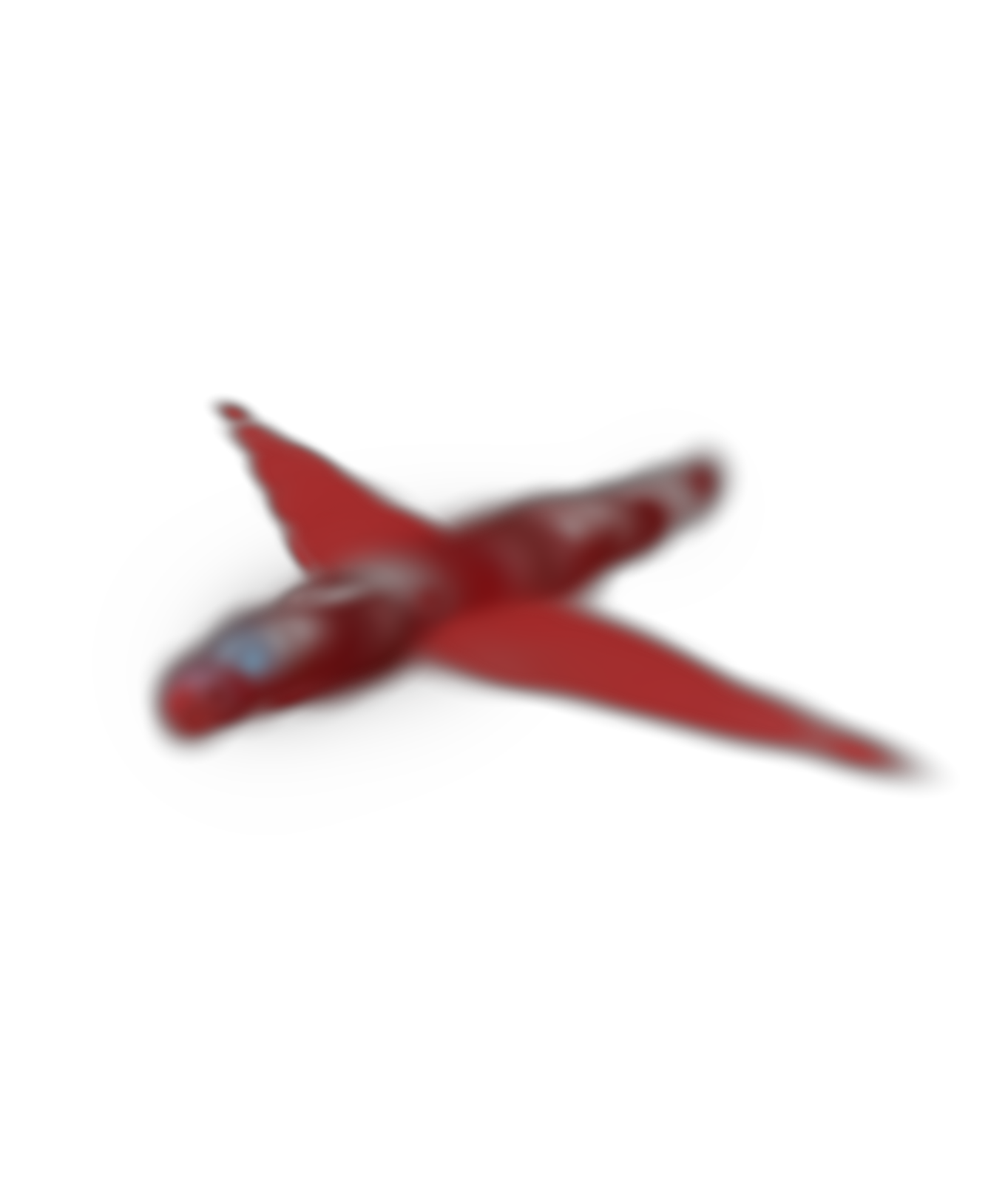} &
        \includegraphics[width=0.24\linewidth]{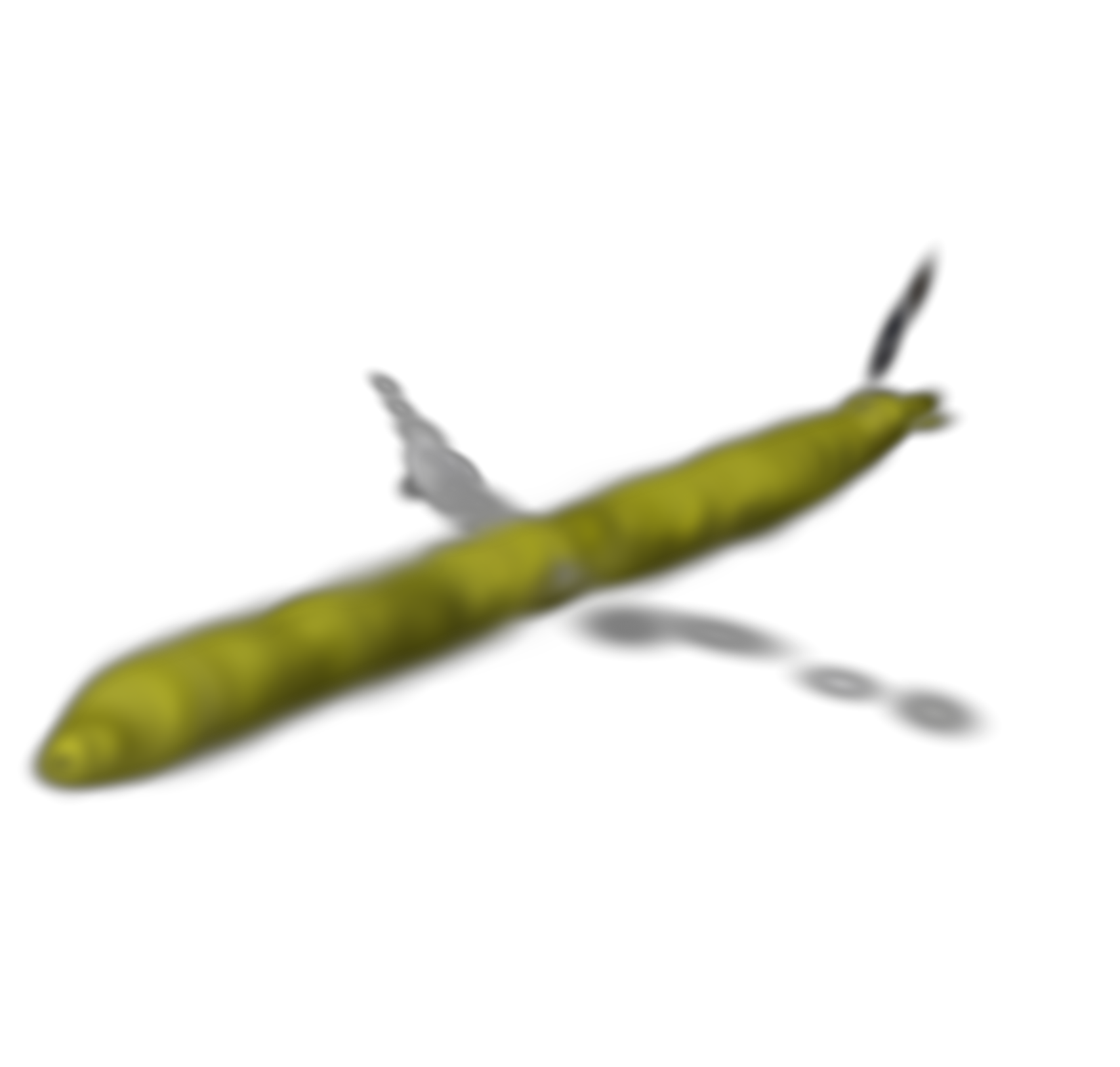} 
    \end{tabular}
    \caption{unconditional generation of in domain airplanes}
    \label{fig:aircraft_tabular}
\end{figure}

To comprehensively evaluate the generative capability of our framework, we conduct a single-class generation experiment, where the model is trained and tested on objects from the same semantic category. We focus on the representative category ``Airplane'' using data from the ShapeSplat dataset. Specifically, the model is trained on ShapeSplat and evaluated for in-domain generation. During training, our model achieves 99.60\% accuracy in detecting whether a node is splittable and 91.45\% accuracy in predicting the corresponding Gaussian properties. Quantitative reconstruction comparisons with the ground truth are reported in Fig.~\ref{fig:quality_curves}, the progressive generation process is illustrated in Fig.~\ref{fig:progressive_gen_single}, and representative unconditional generations are shown in Fig.~\ref{fig:aircraft_tabular}.

\subsection{Applications}

The hierarchical and autoregressive nature of ARGS enables a range of practical applications beyond mere reconstruction. By explicitly modeling multi-scale Gaussian structures, ARGS supports both efficient data handling and flexible content manipulation.

\noindent \textbf{Scene Compression and Transmission.} The reversible Gaussian Splatting reduction naturally functions as a compression mechanism. Each elimination step encodes how fine-scale information can be reconstructed from a compact parent representation. Compared to mesh- or voxel-based codecs, ARGS provides a continuous, differentiable, and reconstruction-aware compression scheme. The hierarchical structure also enables progressive decoding, akin to scalable image codecs, facilitating efficient transmission of large-scale 3D assets or dynamic scenes.

\noindent \textbf{Generation like Large Language Models (LLMs).} Any cut of our sequence can also decode a scene. Similar to how LLMs generate coherent and semantically complete text from arbitrary prefixes, our hierarchical sequence retains structural integrity at every truncation point. Each segment of the sequence—whether terminated at an early coarse level or a later fine-grained level—encapsulates sufficient geometric and appearance priors to reconstruct a meaningful scene representation. This property enables flexible, on-demand generation: users can decode a low-detail scene preview from a short sequence cut for fast visualization, or extend the sequence to finer levels, analogous to LMs generating concise summaries or detailed essays from the same initial prompt. 
% Moreover, this ``prefix-decodable" characteristic inherits the autoregressive essence of LLMs, where each level of the sequence builds upon prior context to maintain consistency, ensuring that even partial sequences produce logically coherent and visually plausible 3D content.

\noindent \textbf{Controllable and Editable Generation.}
Each Gaussian node in ARGS explicitly represents a local spatial structure, enabling semantic or spatial manipulation at multiple levels. Users can modify coarse-level nodes to adjust global shape or pose, or edit subtrees to alter textures or local geometry.

% \begin{figure}[htbp]
%     \centering
%     \setlength{\tabcolsep}{2pt}
%     \renewcommand{\arraystretch}{0}
%     \begin{tabular}{ccc}
%         \includegraphics[width=0.28\linewidth]{figure/airplane/out_domain_1.png} &
%         \includegraphics[width=0.28\linewidth]{figure/airplane/out_domain_2.png} &
%         \includegraphics[width=0.28\linewidth]{figure/airplane/out_domain_3.png} \\
%     \end{tabular}
%     \caption{unconditional generation of out domain airplanes}
%     \label{fig:aircraft_tabular}
% \end{figure}

\subsection{Limitations and Future Works}

Although ARGS demonstrates strong hierarchical modeling, several limitations remain that suggest avenues for future improvement.

\noindent \textbf{Conditional Generation.} Our current model does not support conditional generation, meaning it can only decode objects by category. Future work could incorporate conditions from images or text via cross-attention in the attention blocks.

\noindent \textbf{Token Compactness.} We represent each token as a single Gaussian. However, this representation is inherently inefficient as objects are composed of many Gaussians. To reduce sequence length, it is desirable to use multiple Gaussians per token, thereby achieving a more compact representation.

\begin{figure}[t]
    \centering
    \setlength{\tabcolsep}{2pt}
    \renewcommand{\arraystretch}{0}
    \begin{tabular}{ccc}
        \includegraphics[width=0.32\linewidth]{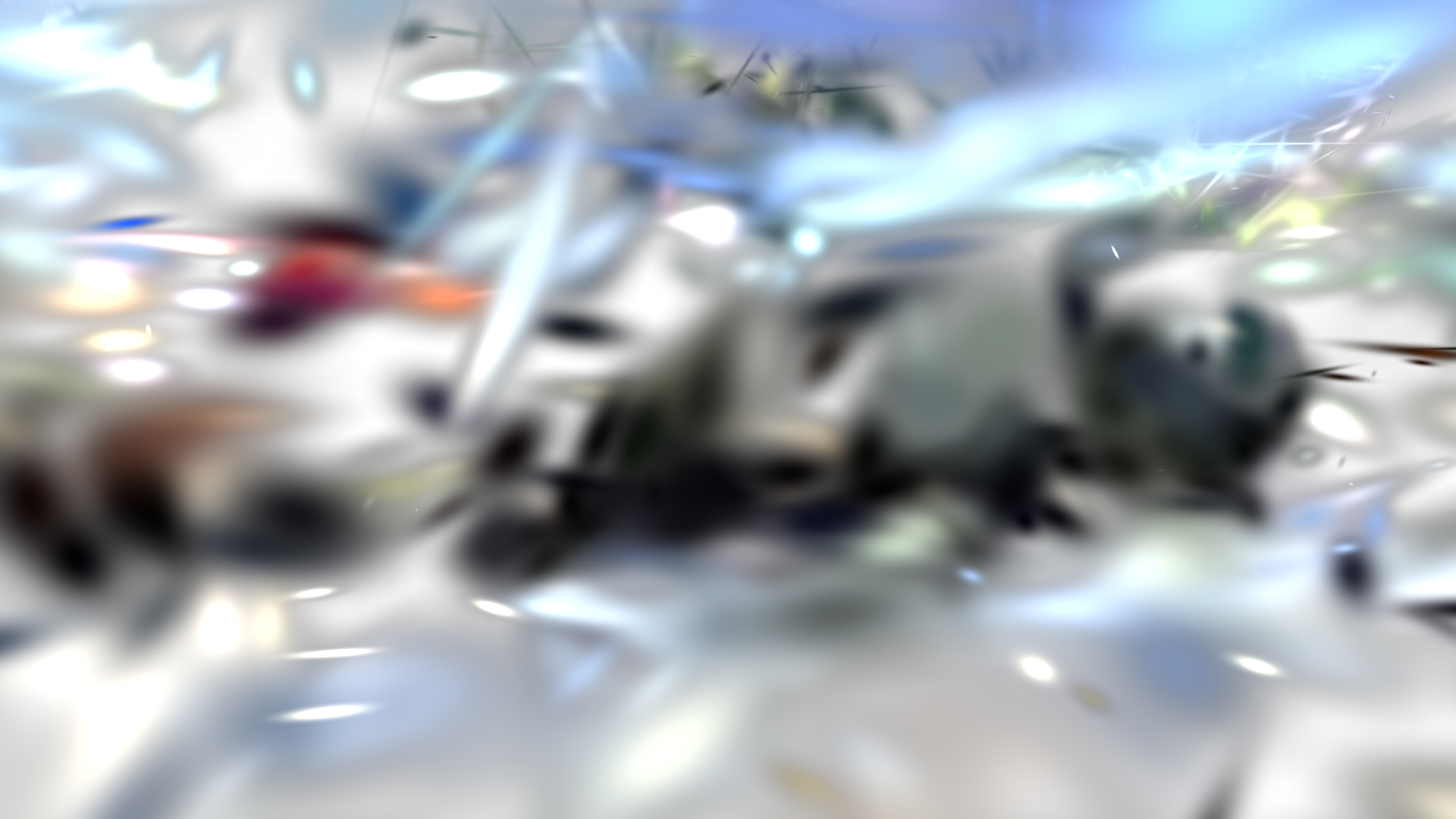}&
        \includegraphics[width=0.32\linewidth]{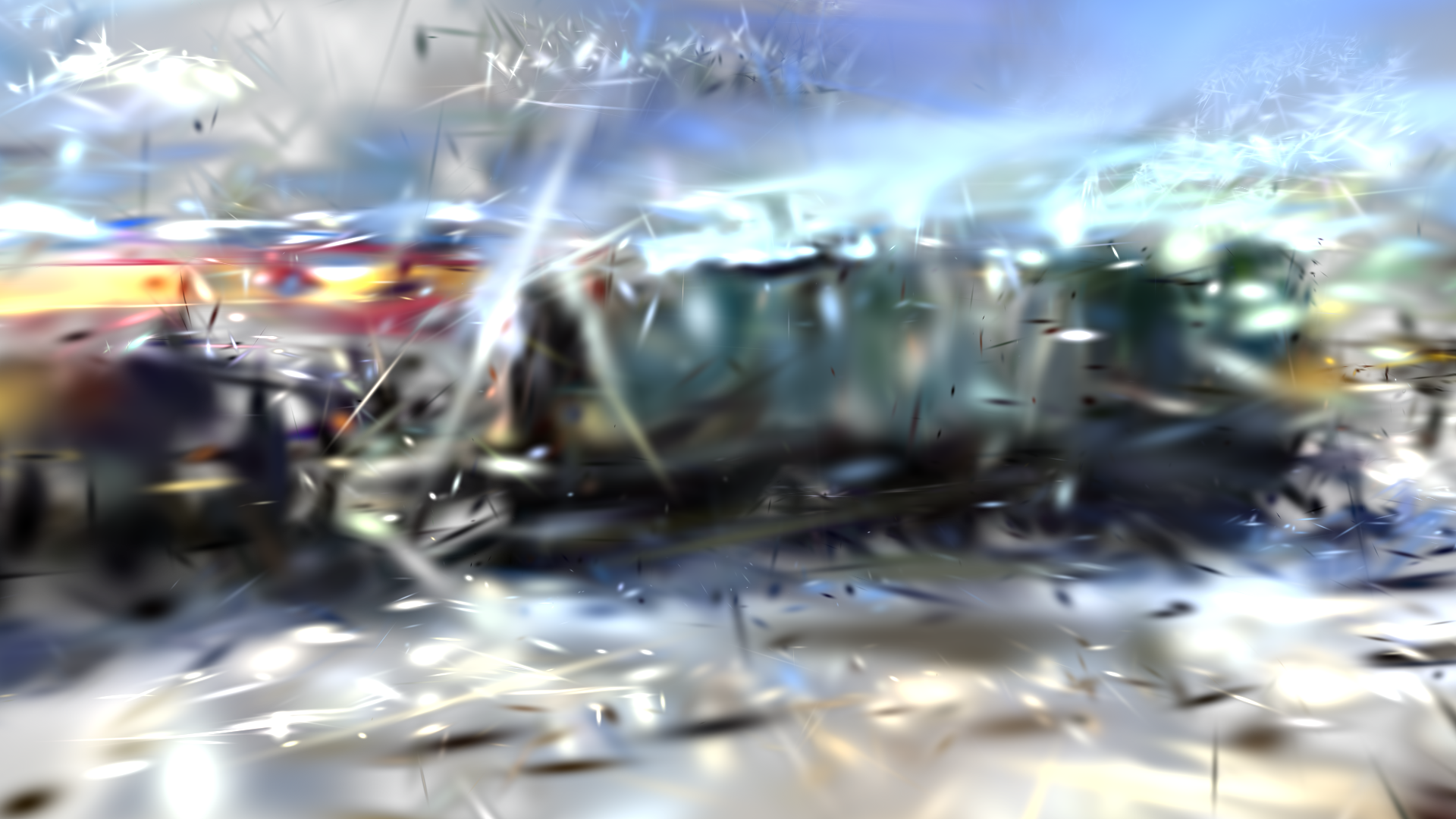}&
        \includegraphics[width=0.32\linewidth]{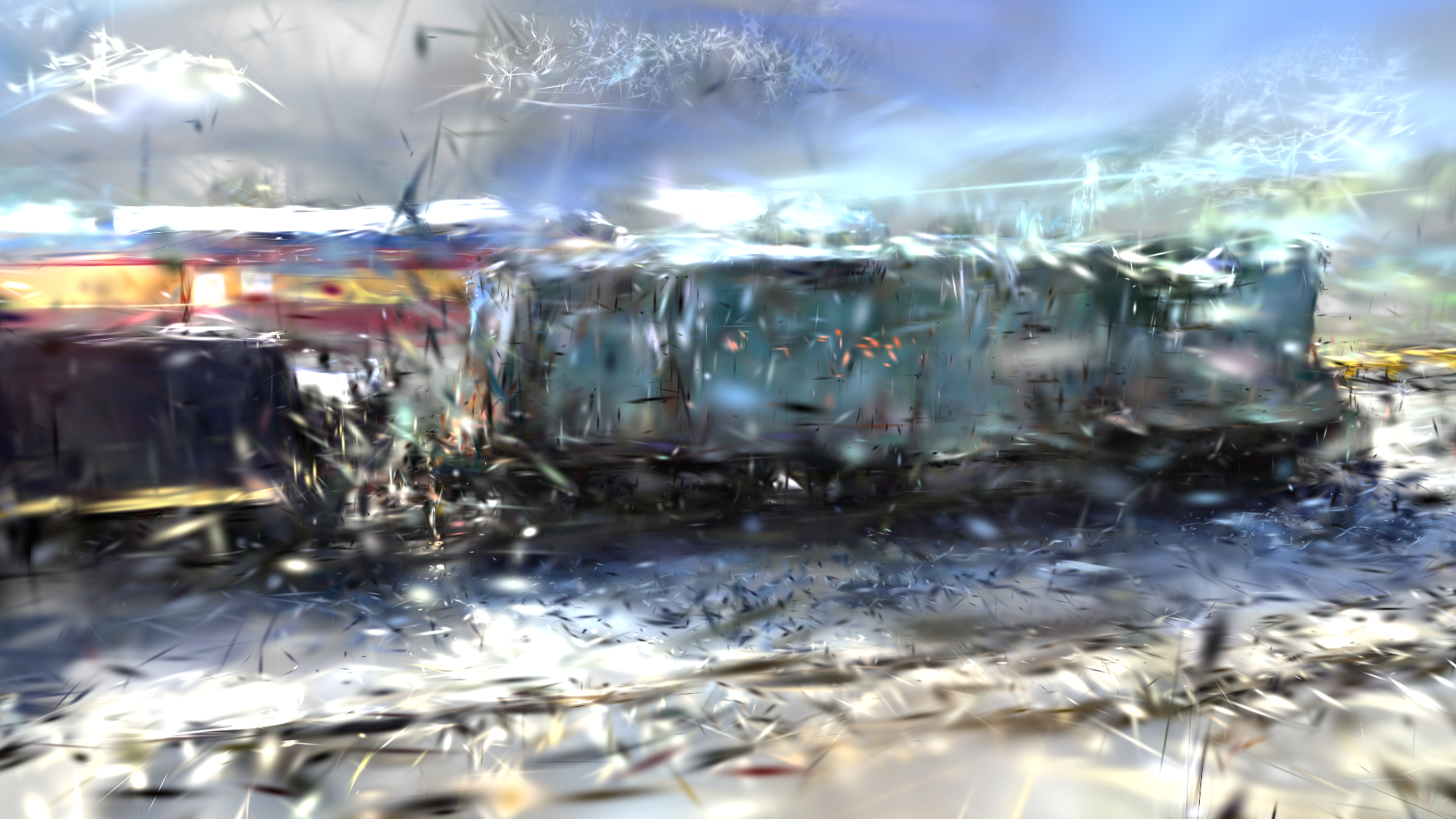}\\

        % ---- 第 2 排 ----
        \includegraphics[width=0.32\linewidth]{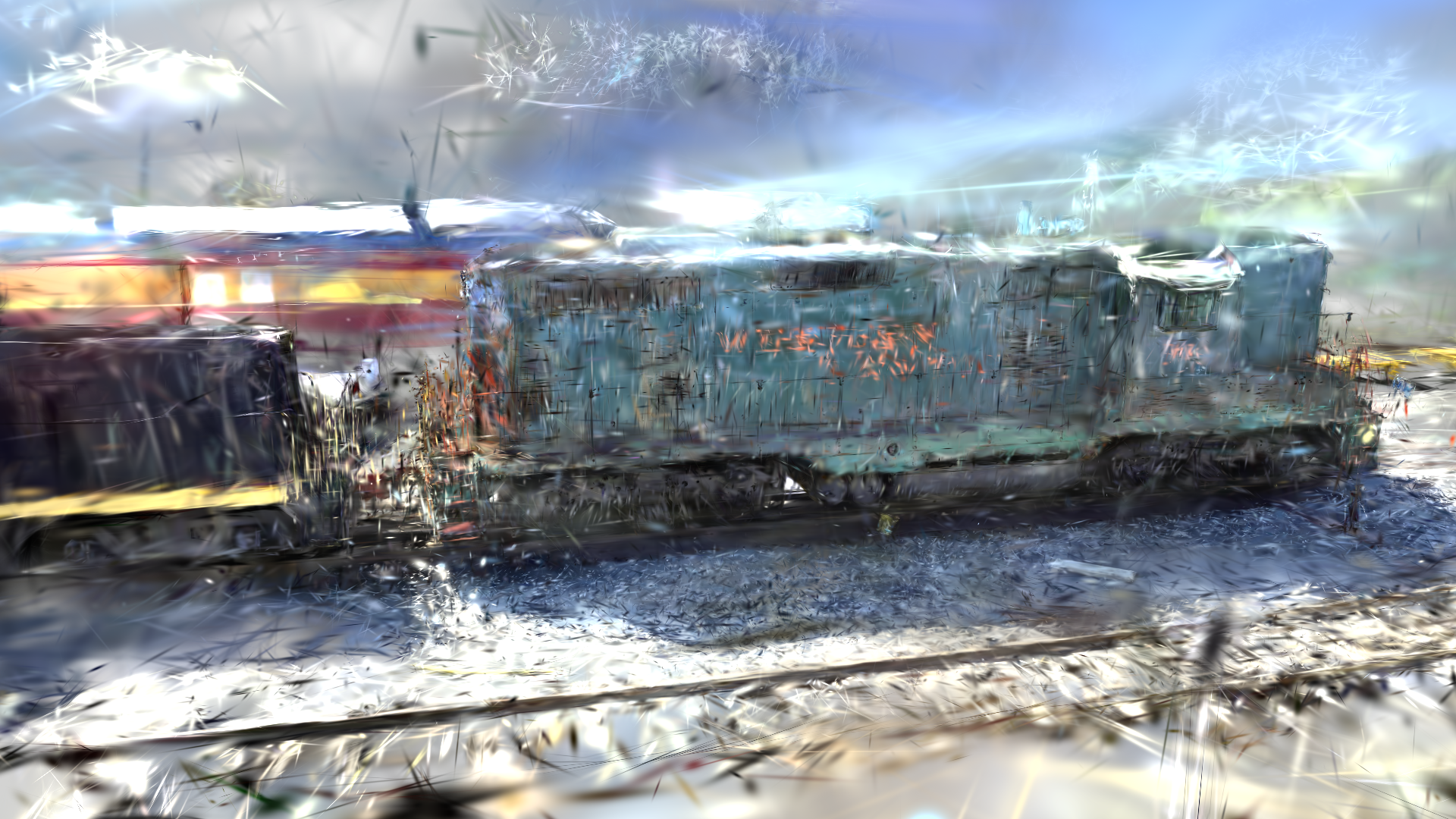}&
        \includegraphics[width=0.32\linewidth]{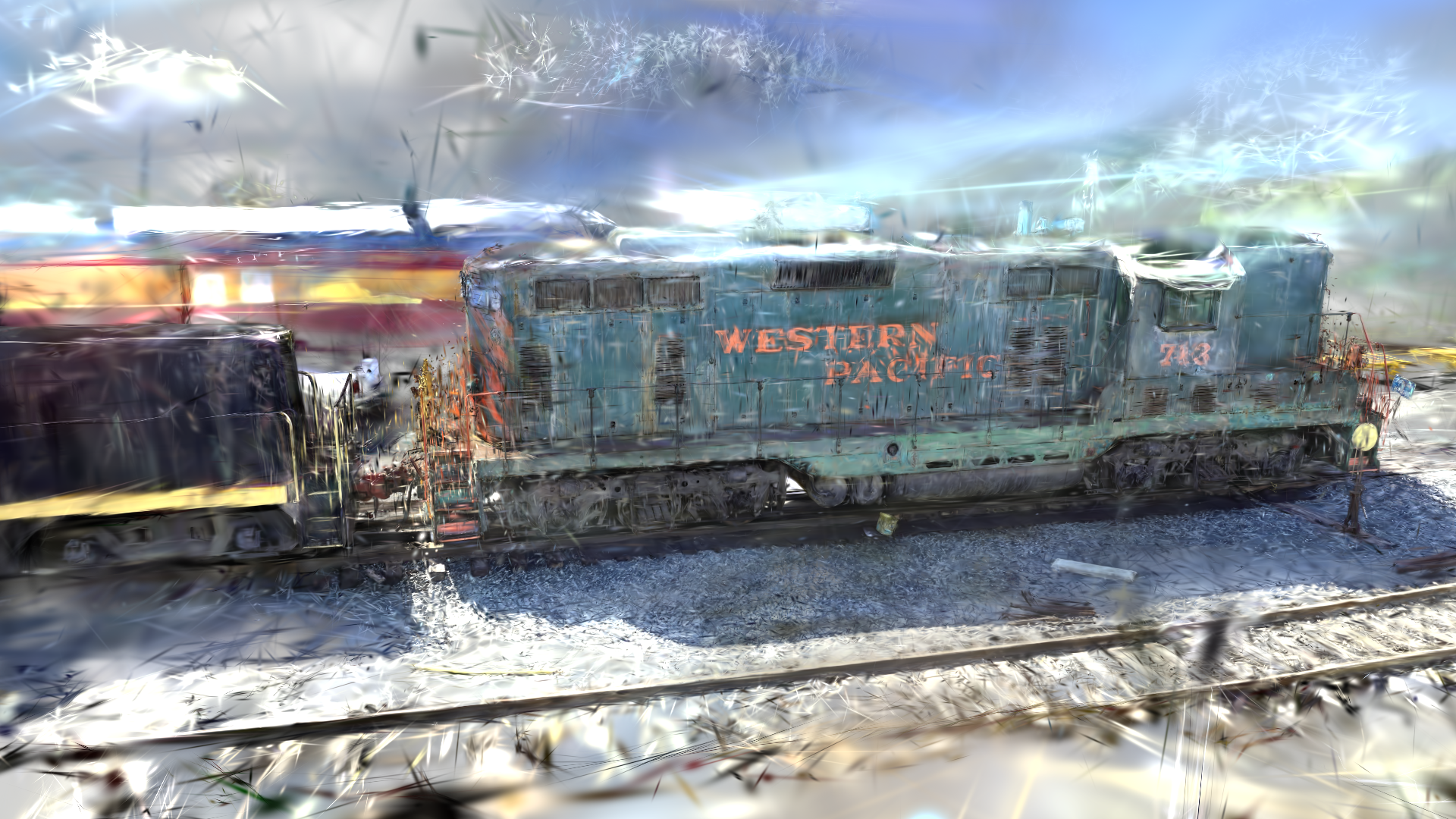}&
        \includegraphics[width=0.32\linewidth]{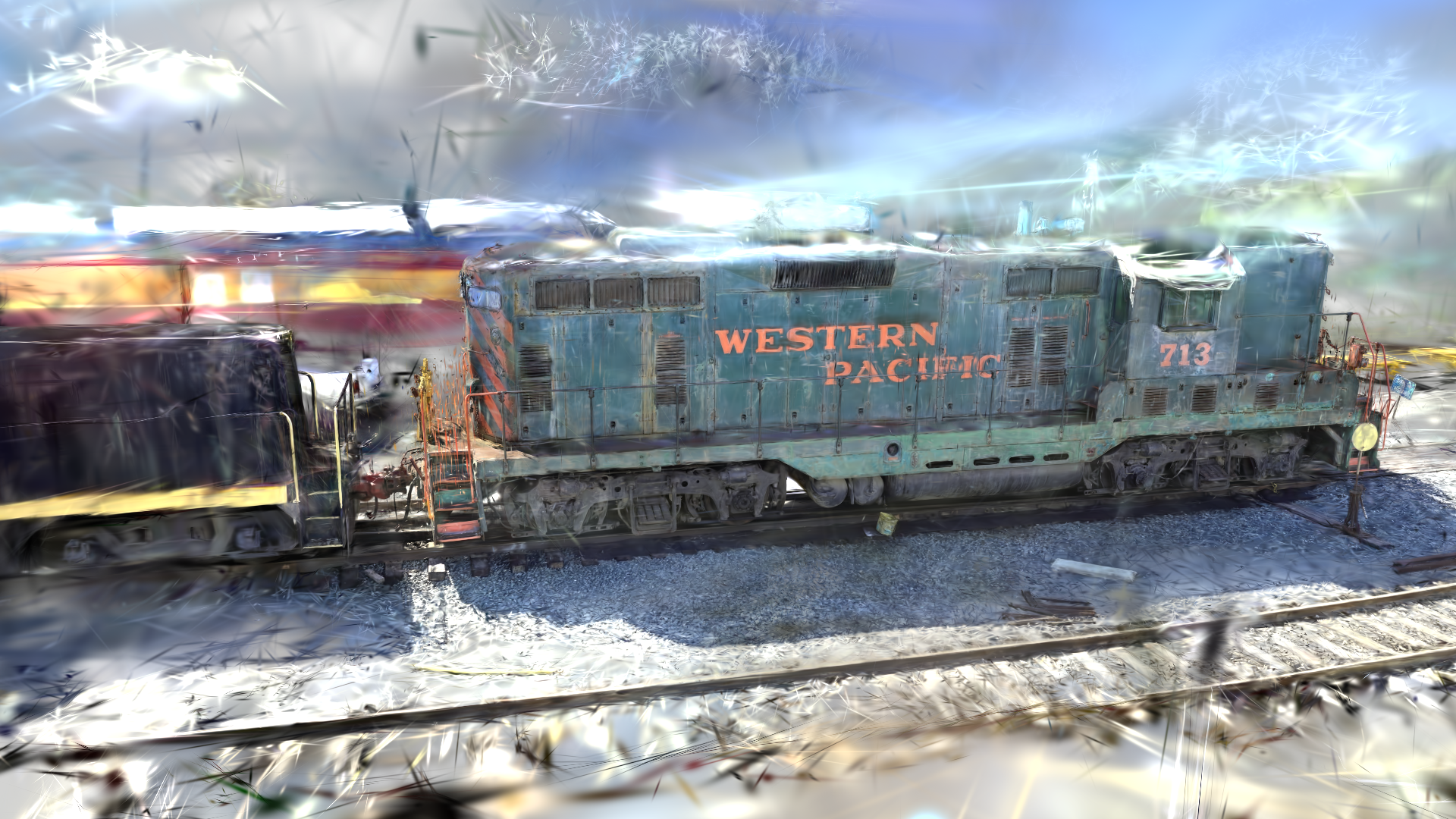}
    \end{tabular}
        \caption{Progressive scene representation from different LoD by our method.}
        \label{fig:lod_halfpage}
\end{figure}

\noindent \textbf{Scene-Level Extension.} Our current assumptions are object-centric rather than scene-centric. In a scene, camera parameters vary, leading to large Gaussians for background regions and small Gaussians for foreground objects. Extending our method to scenes would, therefore, require revised assumptions. Furthermore, our current simplification method has \(O(n^2)\) complexity, as each point requires a neighborhood search over \(n\) points, making it inefficient for large-scale data. A tree-based framework could reduce this to \(O(n \log n)\), enabling efficient processing of large datasets. An example is shown in Fig.~\ref{fig:lod_halfpage}.

\section{Conclusion}
\label{sec:conclusion}

This work proposes a reversible framework for progressive Gaussian simplification and generation, enabling explicit control over reconstruction quality and representation density. It induces a tree-structured hierarchical representation whose complexity grows logarithmically with the number of reduction steps, significantly lowering generation complexity while facilitating multi-scale access and adaptive refinement. Further, a next-scale Gaussian generation method is developed: leveraging hierarchical data from the reduction process, it synthesizes finer-level Gaussian configurations from coarser representations, enabling multi-resolution reconstruction and hierarchical 3D scene synthesis within a unified framework. This method generates Gaussians in sequences analogous to LLMs, and truncating the sequence also allows decoding a scene.
{
    \small
    \bibliographystyle{ieeenat_fullname}
    \bibliography{main}

@String(CVPR= {IEEE Conf. Comput. Vis. Pattern Recog.})

@String(ICCV= {Int. Conf. Comput. Vis.})

@String(ECCV= {Eur. Conf. Comput. Vis.})

@String(CVPR  = {CVPR})

@String(ICCV  = {ICCV})

@String(ECCV  = {ECCV})

@inproceedings{ma2025large,
    title={A Large-Scale Dataset of Gaussian Splats and Their Self-Supervised Pretraining},
    author={Ma, Qi and Li, Yue and Ren, Bin and Sebe, Nicu and Konukoglu, Ender and Gevers, Theo and Van Gool, Luc and Paudel, Danda Pani},
    booktitle={2025 International Conference on 3D Vision (3DV)},
    pages={145--155},
    year={2025},
    organization={IEEE}
}

@inproceedings{kerbl3Dgaussians,
  title={3D Gaussian Splatting for Real-Time Radiance Field Rendering},
  author={Kerbl, Bernhard and Kopanas, Georgios and Leimk{\"u}hler, Thomas and Drettakis, George},
  booktitle={SIGGRAPH},
  year={2023}
}

@article{fang2023splatfields,
  title={SplatFields: Neural Gaussian Splatting Fields for Compact and Precise Surface Reconstruction},
  author={Fang, Jiapeng and others},
  journal={arXiv preprint arXiv:2309.16119},
  year={2023}
}

@inproceedings{yi2023gaussiandreamer,
  title={GaussianDreamer: Fast Generation from Text to 3D Gaussians by Bridging 2D and 3D Diffusion Models},
  author={Yi, Taoran and Fang, Jiemin and Wang, Junjie and Wu, Guanjun and Xie, Lingxi and Zhang, Xiaopeng and Liu, Wenyu and Tian, Qi and Wang, Xinggang},
  year = {2024},
  booktitle = {CVPR}
}

@article{tang2024splatfacto,
  title={Splatfacto: 3D Factorized Gaussians for Compact Representations},
  author={Tang, Jiaxiang and others},
  journal={arXiv preprint arXiv:2403.},
  year={2024}
}

@inproceedings{kazhdan2006poisson,
  title={Poisson surface reconstruction},
  author={Kazhdan, Michael and Bolitho, Matthew and Hoppe, Hugues},
  booktitle={Proceedings of the fourth Eurographics symposium on Geometry processing},
  pages={61--70},
  year={2006}
}

@inproceedings{kato2018renderer,
  title={Neural 3D Mesh Renderer},
  author={Kato, Hiroharu and Ushiku, Yohei and Harada, Tatsuya},
  booktitle={CVPR},
  year={2018}
}

@inproceedings{park2019deepsdf,
  title={DeepSDF: Learning Continuous Signed Distance Functions for Shape Representation},
  author={Park, Jeong Joon and Florence, Peter and Straub, Julian and Newcombe, Richard and Lovegrove, Steven},
  booktitle={CVPR},
  year={2019}
}

@inproceedings{thies2019deferred,
  title={Deferred Neural Rendering: Image Synthesis using Neural Textures},
  author={Thies, Justus and Zollh{\"o}fer, Michael and Stamminger, Marc and Nie{\ss}ner, Matthias},
  booktitle={SIGGRAPH},
  year={2019}
}

@inproceedings{Hoppe1996,
  author    = {Hugues Hoppe},
  title     = {Progressive Meshes},
  booktitle = {Proceedings of SIGGRAPH},
  year      = {1996},
  pages     = {99--108}
}

@inproceedings{Hoppe1997,
  author    = {Hugues Hoppe},
  title     = {View-Dependent Refinement of Progressive Meshes},
  booktitle = {Proceedings of SIGGRAPH},
  year      = {1997},
  pages     = {189--198}
}

@inproceedings{Xia1996,
  author    = {J. C. Xia and Amitabh Varshney},
  title     = {Dynamic View-Dependent Simplification for Polygonal Models},
  booktitle = {Proceedings of IEEE Visualization},
  year      = {1996},
  pages     = {327--334}
}

@inproceedings{Schroeder1992,
  author    = {William J. Schroeder and Jonathan A. Zarge and William E. Lorensen},
  title     = {Decimation of Triangle Meshes},
  booktitle = {Proceedings of SIGGRAPH},
  year      = {1992},
  pages     = {65--70}
}

@inproceedings{Garland1997,
  author    = {Michael Garland and Paul S. Heckbert},
  title     = {Surface Simplification Using Quadric Error Metrics},
  booktitle = {Proceedings of SIGGRAPH},
  year      = {1997},
  pages     = {209--216}
}

@article{barron2022mipnerf360,
    title={Mip-NeRF 360: Unbounded Anti-Aliased Neural Radiance Fields},
    author={Jonathan T. Barron and Ben Mildenhall and 
            Dor Verbin and Pratul P. Srinivasan and Peter Hedman},
    journal={CVPR},
    year={2022}
}

@inproceedings{yu2021plenoctrees,
      title={{PlenOctrees} for Real-time Rendering of Neural Radiance Fields},
      author={Alex Yu and Ruilong Li and Matthew Tancik and Hao Li and Ren Ng and Angjoo Kanazawa},
      year={2021},
      booktitle={ICCV},
}

@article{mueller2022instant,
    author = {Thomas M\"uller and Alex Evans and Christoph Schied and Alexander Keller},
    title = {Instant Neural Graphics Primitives with a Multiresolution Hash Encoding},
    journal = {ACM Trans. Graph.},
    issue_date = {July 2022},
    volume = {41},
    number = {4},
    month = jul,
    year = {2022},
    pages = {102:1--102:15},
    articleno = {102},
    numpages = {15},
    url = {https://doi.org/10.1145/3528223.3530127},
    doi = {10.1145/3528223.3530127},
    publisher = {ACM},
    address = {New York, NY, USA}
}

@INPROCEEDINGS{10670746,
  author={Yu, Zhitao and Yuan, Wei and Zhao, Hengwang and Zhuang, Hanyang and Yang, Ming},
  booktitle={2024 IEEE International Conference on Real-time Computing and Robotics (RCAR)}, 
  title={GOGICP: A Real-time Gaussian Octree-based GICP Method for Faster Point Cloud Registration}, 
  year={2024},
  volume={},
  number={},
  pages={112-117},
  doi={10.1109/RCAR61438.2024.10670746}}

@inproceedings{chen2024sar3d,
    title={SAR3D: Autoregressive 3D Object Generation and Understanding via Multi-scale 3D VQVAE},
    author={Chen, Yongwei and Lan, Yushi and Zhou, Shangchen and Wang, Tengfei and Pan, Xingang},
    booktitle={CVPR},
    year={2025}}

@Article{VAR,
      title={Visual Autoregressive Modeling: Scalable Image Generation via Next-Scale Prediction}, 
      author={Keyu Tian and Yi Jiang and Zehuan Yuan and Bingyue Peng and Liwei Wang},
      year={2024},
      eprint={2404.02905},
      archivePrefix={arXiv},
      primaryClass={cs.CV}
}

@misc{meng2025pointnspautoregressive3dpoint,
      title={PointNSP: Autoregressive 3D Point Cloud Generation with Next-Scale Level-of-Detail Prediction}, 
      author={Ziqiao Meng and Qichao Wang and Zhiyang Dou and Zixing Song and Zhipeng Zhou and Irwin King and Peilin Zhao},
      year={2025},
      eprint={2510.05613},
      archivePrefix={arXiv},
      primaryClass={cs.CV},
      url={https://arxiv.org/abs/2510.05613}, 
}

@misc{lei2025armesh,
      title={ARMesh: Autoregressive Mesh Generation via Next-Level-of-Detail Prediction}, 
      author={Jiabao Lei and Kewei Shi and Zhihao Liang and Kui Jia},
      year={2025},
      eprint={2509.20824},
      archivePrefix={arXiv},
      primaryClass={cs.GR},
      url={https://arxiv.org/abs/2509.20824}, 
}

@InProceedings{Zhang_2025_ICCV_VertexRegen,
      author    = {Zhang, Xiang and Siddiqui, Yawar and Avetisyan, Armen and Xie, Chris and Engel, Jakob and Howard-Jenkins, Henry},
      title     = {VertexRegen: Mesh Generation with Continuous Level of Detail},
      booktitle = {Proceedings of the IEEE/CVF International Conference on Computer Vision (ICCV)},
      month     = {October},
      year      = {2025},
      pages     = {12570-12580}
  }

@article{tang2024lgm,
  title={LGM: Large Multi-View Gaussian Model for High-Resolution 3D Content Creation},
  author={Tang, Jiaxiang and Chen, Zhaoxi and Chen, Xiaokang and Wang, Tengfei and Zeng, Gang and Liu, Ziwei},
  journal={arXiv preprint arXiv:2402.05054},
  year={2024}
}

@inproceedings{
      yang2025atlas,
      title={Atlas Gaussians Diffusion for 3D Generation},
      author={Haitao Yang and Yuan Dong and Hanwen Jiang and Dejia Xu and Georgios Pavlakos and Qixing Huang},
      booktitle={The Thirteenth International Conference on Learning Representations},
      year={2025},
      url={https://openreview.net/forum?id=H2Gxil855b}
    }

@inproceedings{roessle2024l3dg,
      title={L3DG: Latent 3D Gaussian Diffusion}, 
      author={Roessle, Barbara and M{\"u}ller, Norman and Porzi, Lorenzo and Bul{\`o}, Samuel Rota and Kontschieder, Peter and Dai, Angela and Nie{\ss}ner, Matthias},
      booktitle={SIGGRAPH Asia 2024 Conference Papers},
      month={December},
      year={2024}
}

@inproceedings{DiffGS,
      title = {DiffGS: Functional Gaussian Splatting Diffusion},
      author = {Zhou, Junsheng and Zhang, Weiqi and Liu, Yu-Shen},
      booktitle = {Advances in Neural Information Processing Systems (NeurIPS)},
      year = {2024}
  }

@inproceedings{Infinity,
  title={Infinity: Scaling bitwise autoregressive modeling for high-resolution image synthesis},
  author={Han, Jian and Liu, Jinlai and Jiang, Yi and Yan, Bin and Zhang, Yuqi and Yuan, Zehuan and Peng, Bingyue and Liu, Xiaobing},
  booktitle={Proceedings of the Computer Vision and Pattern Recognition Conference},
  pages={15733--15744},
  year={2025}
}

@inproceedings{mildenhall2020nerf,
  title={NeRF: Representing Scenes as Neural Radiance Fields for View Synthesis},
  author={Ben Mildenhall and Pratul P. Srinivasan and Matthew Tancik and Jonathan T. Barron and Ravi Ramamoorthi and Ren Ng},
  year={2020},
  booktitle={ECCV},
}

@article{liu2019softras,
  title={Soft Rasterizer: A Differentiable Renderer for Image-based 3D Reasoning},
  author={Liu, Shichen and Li, Tianye and Chen, Weikai and Li, Hao},
  journal={The IEEE International Conference on Computer Vision (ICCV)},
  month = {Oct},
  year={2019}
}

@article{liu2020general,
  title={A General Differentiable Mesh Renderer for Image-based 3D Reasoning},
  author={Liu, Shichen and Li, Tianye and Chen, Weikai and Li, Hao},
  journal={IEEE Transactions on Pattern Analysis and Machine Intelligence},
  year={2020},
  publisher={IEEE}
}

@article{ho2020denoising,
    title={Denoising Diffusion Probabilistic Models},
    author={Jonathan Ho and Ajay Jain and Pieter Abbeel},
    year={2020},
    journal={arXiv preprint arxiv:2006.11239}
}

@techreport{shapenet2015,
  title       = {{ShapeNet: An Information-Rich 3D Model Repository}},
  author      = {Chang, Angel X. and Funkhouser, Thomas and Guibas, Leonidas and Hanrahan, Pat and Huang, Qixing and Li, Zimo and Savarese, Silvio and Savva, Manolis and Song, Shuran and Su, Hao and Xiao, Jianxiong and Yi, Li and Yu, Fisher},
  number      = {arXiv:1512.03012 [cs.GR]},
  institution = {Stanford University --- Princeton University --- Toyota Technological Institute at Chicago},
  year        = {2015}
}

@inproceedings{wu20153d,
  title={3d shapenets: A deep representation for volumetric shapes},
  author={Wu, Zhirong and Song, Shuran and Khosla, Aditya and Yu, Fisher and Zhang, Linguang and Tang, Xiaoou and Xiao, Jianxiong},
  booktitle={Proceedings of the IEEE conference on computer vision and pattern recognition},
  pages={1912--1920},
  year={2015}
}

@article{objaverse,
  title={Objaverse: A Universe of Annotated 3D Objects},
  author={Matt Deitke and Dustin Schwenk and Jordi Salvador and Luca Weihs and
          Oscar Michel and Eli VanderBilt and Ludwig Schmidt and
          Kiana Ehsani and Aniruddha Kembhavi and Ali Farhadi},
  journal={arXiv preprint arXiv:2212.08051},
  year={2022}
}
}


\end{document}